\documentclass{article}

\usepackage{arxiv}
\usepackage{natbib}

\usepackage[utf8]{inputenc}
\usepackage[T1]{fontenc}
\usepackage{url}
\usepackage{booktabs}
\usepackage{amsfonts}
\usepackage{nicefrac}
\usepackage{microtype}
\usepackage{xcolor}
\usepackage{graphicx}
\usepackage{amsmath}
\usepackage{multirow}
\usepackage{float}
\usepackage[hidelinks]{hyperref}
 
\def\dagger{\text{\textsuperscript{+}}}
\def\ddagger{\text{\textsuperscript{--}}}

\title{VANDAM: viewing a nucleotide sequence with DNA molecular priors}

\author{
\parbox{0.97\textwidth}{\centering
\mbox{\href{https://orcid.org/0000-0001-7498-6413}{\includegraphics[scale=0.06]{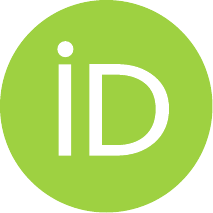}\hspace{1mm}Jeremy Levy$^{1}$}}
\quad
\mbox{\href{https://orcid.org/0000-0002-5006-9300}{\includegraphics[scale=0.06]{orcid.pdf}\hspace{1mm}Ariel Larey$^{1}$}}
\quad
\mbox{\href{https://orcid.org/0009-0007-3657-6270}{\includegraphics[scale=0.06]{orcid.pdf}\hspace{1mm}Yury Nahshan$^{1}$}}
\quad
\mbox{\href{https://orcid.org/0009-0000-0029-342X}{\includegraphics[scale=0.06]{orcid.pdf}\hspace{1mm}Raizy Kellerman$^{1}$}}
\quad
\mbox{\href{https://orcid.org/0009-0009-5635-6763}{\includegraphics[scale=0.06]{orcid.pdf}\hspace{1mm}Elay Dahan$^{2}$}}
\quad
\mbox{\href{https://orcid.org/0009-0004-8952-3672}{\includegraphics[scale=0.06]{orcid.pdf}\hspace{1mm}Amit Bleiweiss$^{3}$}}
\quad
\mbox{\href{https://orcid.org/0009-0007-7314-7934}{\includegraphics[scale=0.06]{orcid.pdf}\hspace{1mm}Guy Leib$^{4}$}}
\quad
\mbox{\href{https://orcid.org/0000-0001-5060-891X}{\includegraphics[scale=0.06]{orcid.pdf}\hspace{1mm}Omri Nayshool$^{4}$}}
\quad
\mbox{\href{https://orcid.org/0000-0001-5136-8014}{\includegraphics[scale=0.06]{orcid.pdf}\hspace{1mm}Dan Ofer$^{4}$}}
\quad
\mbox{\href{https://orcid.org/0000-0002-5580-5282}{\includegraphics[scale=0.06]{orcid.pdf}\hspace{1mm}Tal Zinger$^{4}$}}
\quad
\mbox{\href{https://orcid.org/0000-0003-1121-5130}{\includegraphics[scale=0.06]{orcid.pdf}\hspace{1mm}Dan Dominissini$^{4}$}}
\quad
\mbox{\href{https://orcid.org/0000-0002-4594-1811}{\includegraphics[scale=0.06]{orcid.pdf}\hspace{1mm}Gideon Rechavi$^{4, 5}$}}
\quad
\mbox{\href{https://orcid.org/0000-0001-6488-6240}{\includegraphics[scale=0.06]{orcid.pdf}\hspace{1mm}Marissa Wirth$^{6}$}}
\quad
\mbox{\href{https://orcid.org/0000-0003-1874-2515}{\includegraphics[scale=0.06]{orcid.pdf}\hspace{1mm}Simon Lee$^{6}$}}
\quad
\mbox{\href{https://orcid.org/0000-0001-5951-1866}{\includegraphics[scale=0.06]{orcid.pdf}\hspace{1mm}Dung Hoang$^{6}$}}
\quad
\mbox{\href{https://orcid.org/0000-0002-4258-1574}{\includegraphics[scale=0.06]{orcid.pdf}\hspace{1mm}Noam D. Beckmann$^{6}$}}
\quad
\mbox{\href{https://orcid.org/0000-0002-1693-9157}{\includegraphics[scale=0.06]{orcid.pdf}\hspace{1mm}Shane O'Connell$^{6}$}}
\quad
\mbox{\href{https://orcid.org/0000-0002-2609-4354}{\includegraphics[scale=0.06]{orcid.pdf}\hspace{1mm}Nicole Bussola$^{6}$}}
\quad
\mbox{\href{https://orcid.org/0000-0001-8135-6858}{\includegraphics[scale=0.06]{orcid.pdf}\hspace{1mm}Alexander W. Charney$^{6}$}}
\quad
\mbox{\href{https://orcid.org/0000-0002-3393-6070}{\includegraphics[scale=0.06]{orcid.pdf}\hspace{1mm}Yoli Shavit$^{1,\ast}$}}
\quad
\mbox{\href{https://orcid.org/0000-0002-0939-3379}{\includegraphics[scale=0.06]{orcid.pdf}\hspace{1mm}Nati Daniel$^{1,\ast}$}}
\endgraf\vspace{0.75em}
{\normalfont\footnotesize
$^{1}$Applied AI Architecture, NVIDIA, Israel.\\
$^{2}$Worldwide Field Ops, NVIDIA, Israel.\\
$^{3}$Developer Programs, NVIDIA, Israel.\\
$^{4}$Cancer Research Center and Wohl Institute of Translational Medicine, Sheba Medical Center, Tel Hashomer, Israel.\\
$^{5}$Dina Recanati School of Medicine Reichman University, Israel. \\
$^{6}$Windreich Department of AI and Human Health, Icahn School of Medicine at Mount Sinai, New York, USA.
\endgraf\vspace{0.5em}
$^{\ast}$Co-corresponding authors: \href{mailto:ndaniel@nvidia.com}{ndaniel@nvidia.com},
\href{mailto:yolis@nvidia.com}{yolis@nvidia.com}.
\endgraf}
}}

\date{}
\renewcommand{\shorttitle}{VANDAM: viewing a nucleotide sequence with DNA molecular priors}
\hypersetup{
  pdftitle={VANDAM: viewing a nucleotide sequence with DNA molecular priors},
  pdfauthor={Jeremy Levy, Ariel Larey, Yury Nahshan, Raizy Kellerman, Elay Dahan, Amit Bleiweiss, Guy Leib, Omri Nayshool, Dan Ofer, Tal Zinger, Dan Dominissini, Gideon Rechavi, Marissa Wirth, Simon Lee, Dung Hoang, Noam D. Beckmann, Shane O'Connell, Nicole Bussola, Alexander W. Charney, Yoli Shavit, Nati Daniel}
}

\begin{document}

\maketitle

%%%%%%%%%%%%%%%%%%%%%%%%%%%%%%%%%%%%%%%%%%%%%%%%%%%%%%%%%%%%
\begin{abstract}
Contemporary Genomic Foundation Models (GFMs) rely on a DNA-as-a-string paradigm that employs masked token prediction objectives for pretraining. However, this abstraction does not explicitly model the biochemical, structural, and physical properties essential to biological function. Many molecular properties can be estimated from sequence using established biophysical models, so their utility lies not in providing an independent modality, but in introducing priors that training objectives can explicitly exploit.
We introduce VANDAM, a framework that extends the training of GFMs with DNA molecular priors. In self-supervised training, VANDAM predicts regional molecular properties from pooled representations. When functional labels are available and can reward retaining molecular priors, local features are additionally injected at the input.
VANDAM consistently improves downstream performance across four architecture families and  nine held-out genomic tasks by complementing token-based objectives. Probing experiments further demonstrate that the use of molecular priors generalizes to other unseen molecular properties.
\end{abstract}

%%%%%%%%%%%%%%%%%%%%%%%%%%%%%%%%%%%%%%%%%%%%%%%%%%%%%%%%%%%%
\section{Introduction}
%%%%%%%%%%%%%%%%%%%%%%%%%%%%%%%%%%%%%%%%%%%%%%%%%%%%%%%%%%%%

Genomic foundation models (GFMs) learn transferable representations from large unlabeled genomic corpora and support tasks ranging from regulatory element annotation to variant effect prediction \citep{benegas2025genomic, zhou2024dnabert, boshar2025foundational, schiff2024caduceus, brixi2026genome}.
Despite substantial diversity in architecture, scale, and context length, their training objectives remain largely inherited from natural language processing (NLP), where masked language modeling (MLM) and next-token prediction (NTP) train models to recover or predict token identity, treating nucleotides primarily as discrete symbols. In NLP, pre-training and many downstream applications share a common linguistic interface, as the model conditions on text to produce text. For generative tasks such as question answering, NTP directly trains the mechanism used to generate an answer. Genomic downstream tasks, by contrast, predict functional and phenotypic properties, such as regulatory activity, gene expression, and pathogenicity. Token prediction remains a useful general-purpose prior, but it does not directly require the representation to preserve the molecular properties relevant to these tasks.

DNA differs from natural language in an important respect: whereas linguistic semantics is inferred from usage, many molecular properties of DNA have been characterized and can be computed directly from sequence. For example, nearest-neighbor models provide duplex free-energy contributions \citep{santalucia1998unified}, and helical shape parameters can be estimated using models calibrated against all-atom simulations \citep{zhou2013dnashape, chiu2016dnashaper, li2024predicting}. We therefore argue that GFMs should represent DNA not merely as a string of symbols, but also as a molecule with predictable structural, energetic, and chemical properties.

Motivated by this distinction and by the successful use of structural and functional priors in protein models, we introduce VANDAM, a framework that complements the conventional DNA-as-a-string view with a DNA-as-a-molecule view. VANDAM incorporates molecular priors through two complementary mechanisms. Local Feature Injection (LFI) augments token embeddings with local molecular descriptors, while Region Property Prediction (RPP) adds an auxiliary objective that asks pooled representations to predict sequence-derived molecular properties of genomic regions. RPP is employed in self-supervised pre-training, while LFI is additionally used in multi-task training, where functional labels can reward the use of molecular descriptors. Together, these mechanisms provide input-level and representation-level routes for grounding GFMs in the physical properties of DNA.

Across four backbone families and nine held-out tasks, VANDAM improves the point estimate in 29 of 36 compute-matched pre-training comparisons and 33 of 36 multi-task comparisons; 25 and 23 comparisons, respectively, show statistical significance. Matched non-biological controls show that these gains depend on biological target content rather than auxiliary capacity alone, and RPP increases linear access to six of eight held-out qDNA descriptors on an independent confirmation split.

We make four contributions. First, we frame genomic representation learning around objective-appropriate molecular grounding, treating DNA as a molecule. Second, we introduce LFI and RPP, which respectively inject local descriptors when functional labels can reward their use and directly supervise regional molecular properties during self-supervised training. Third, we evaluate VANDAM through compute-matched experiments across four backbone families and nine held-out tasks, and use matched non-biological controls to distinguish biological content from auxiliary capacity and optimization. Finally, we show with frozen-representation probes that RPP increases access to an independent family of held-out molecular descriptors.
%%%%%%%%%%%%%%%%%%%%%%%%%%%%%%%%%%%%%%%%%%%%%%%%%%%%%%%%%%%%
\section{Related Work}
%%%%%%%%%%%%%%%%%%%%%%%%%%%%%%%%%%%%%%%%%%%%%%%%%%%%%%%%%%%%

\paragraph{Genomic foundation models.}
\label{sec:rw-gfm}
Genomic foundation models (GFMs) follow the paradigm established by large language models: pre-train a sequence model on unlabeled genomic corpora, then transfer it to downstream functional genomics tasks.
The landscape comprises diverse architectures and tokenization schemes for genomic sequence modeling.
DNABERT-2 \citep{zhou2024dnabert} keeps a flat transformer encoder but replaces the overlapping $k$-mer tokenization of its predecessor \citep{ji2021dnabert} 
with byte-pair encoding; NTv3 \citep{boshar2025foundational} adopts a U-Net-style convolutional--transformer hybrid; Caduceus \citep{schiff2024caduceus} 
replaces attention with a bi-directional Mamba block and, in its parameter-sharing variant, enforces reverse-complement equivariance architecturally; and Evo2 \citep{brixi2026genome} scales a StripedHyena2 hybrid to 40B parameters 
trained on 9.3 trillion tokens.
Despite this diversity, most prominent GFMs rely primarily on pre-training objectives inherited from natural language processing: masked recovery or autoregressive prediction of token identity.
Whether a token denotes a single nucleotide or a byte-pair-encoded sequence fragment, the objective treats it as a discrete symbol and does not explicitly supervise the physical properties of the underlying DNA sequence.

This token-identity objective has recently been questioned, on the grounds that it captures local genomic syntax and motif structure while leaving broader functional context underdetermined.
Several recent approaches therefore modify the training objective or the genomic representation: JEPA-DNA \citep{larey2026jepa} adds supervision of high-level functional embeddings of masked segments alongside token-level reconstruction; D3LM \citep{yang2026d3lm} reframes DNA modeling as discrete diffusion; UKBioBERT \citep{liu2026pre} continues masked-language modeling on sequences containing population-level genetic variation rather than reference sequence alone; and BioToken/BioFM \citep{medvedev2025biotoken} encodes variants and genomic annotations as biologically informed tokens.
DNAGPT \citep{zhang2023dnagpt} augments autoregressive pre-training with sequence-order classification and GC-content regression over sampled sequence regions.
MutBERT \citep{long2025mutbert} represents each position as a population allele-frequency distribution; GPN-MSA \citep{benegas2025dna} uses multi-species alignments as input, and PhyloGPN \citep{albors2025phylogenetic} incorporates phylogenetic alignments through its training objective while retaining single-sequence inference. These results suggest that the training signal and the representation of genomic context, not only capacity or architecture, affect the quality of learned representations.

\paragraph{Biophysical supervision in representation learning.}
\label{sec:rw-injection}
A relevant precedent is UTR-LM \citep{chu20245}, which augments masked-nucleotide modeling of 5$'$ untranslated regions with auxiliary targets computed from the same sequence by ViennaRNA: secondary structure at masked positions and sequence-level minimum free energy.
These targets provide biophysical supervision without introducing an additional experimental or structural modality.
In DNA modeling, MIXBend \citep{yang2024employing} aligns DNABERT sequence representations with physicochemical features during self-supervised pre-training, but develops this approach specifically for DNA bendability prediction.
Whereas UTR-LM focuses on RNA and MIXBend is tailored to DNA bendability, VANDAM develops a general framework for grounding existing GFMs in a broader set of sequence-derived DNA structural, energetic, and chemical properties under both self-supervised and functionally supervised training.

\paragraph{Biophysical features in supervised prediction.}
\label{sec:rw-dnashape}
Biophysical features carry well-established predictive signal for DNA, but much of the existing evidence comes from task-specific supervised models.
DNA-shape features such as minor-groove width, computed from pentamer lookup tables built from all-atom Monte Carlo simulations
\citep{zhou2013dnashape,chiu2016dnashaper,li2017expanding},
improve transcription-factor-binding-site prediction over position-weight matrices and sequence-only baselines
\citep{zhou2015quantitative,mathelier2016dna}.
More recent work replaces the lookup tables with learned predictors that account for extended flanking context
\citep{li2024predicting}.
DNABERT-CoreProm-MM \citep{papineni2025augmenting} augments DNABERT embeddings with nucleotide composition, CpG density, minor-groove width, electrostatic potential, and other DNA-intrinsic features for supervised core-promoter prediction.
Together, these studies establish the utility of sequence-derived molecular descriptors, but primarily within task-specific supervised settings.
%%%%%%%%%%%%%%%%%%%%%%%%%%%%%%%%%%%%%%%%%%%%%%%%%%%%%%%%%%%%
\section{Methods}
\label{sec:methods}
%%%%%%%%%%%%%%%%%%%%%%%%%%%%%%%%%%%%%%%%%%%%%%%%%%%%%%%%%%%%

\subsection{From DNA as a String to DNA as a Molecule}
\label{sec:molecular-view}

\begin{figure}[t]
\centering
\includegraphics[width=\textwidth]{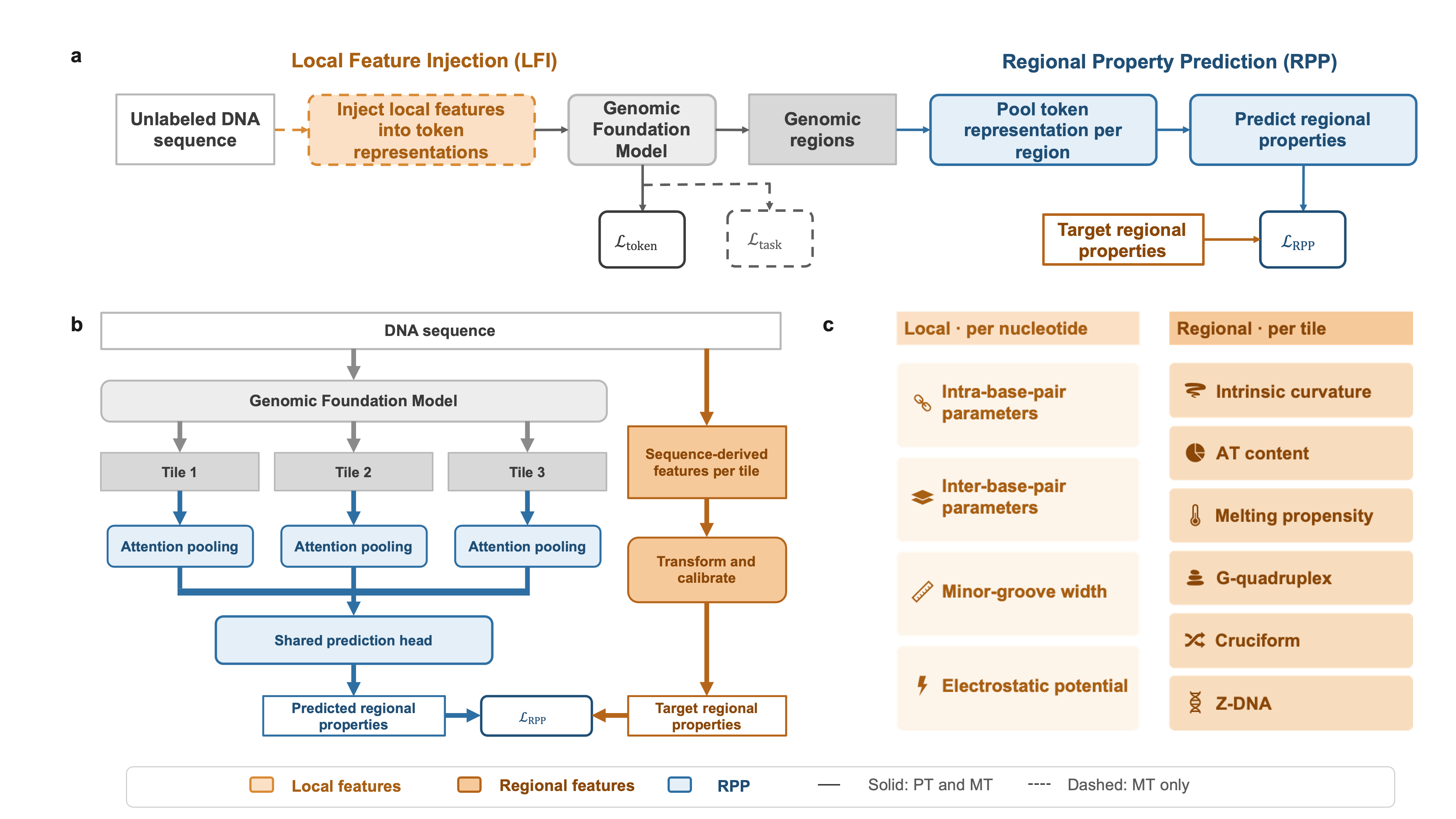}
\caption{VANDAM architecture and molecular feature hierarchy. 
\textbf{a}, High-level training flow. Local Feature Injection (LFI; dashed) augments input token representations with local sequence-derived descriptors and is used only in VANDAM-MT, where functional-task supervision $\mathcal{L}_{\mathrm{task}}$ can reward their use. 
Regional Property Prediction (RPP; blue) pools representations for genomic regions and predicts tile-level properties in both VANDAM-PT and VANDAM-MT, alongside the native token objective $\mathcal{L}_{\mathrm{token}}$. 
\textbf{b}, RPP mechanism. Backbone representations are partitioned into tiles, attention-pooled within each tile, passed through a shared prediction head, and compared with transformed and calibrated sequence-derived regional targets using $\mathcal{L}_{\mathrm{RPP}}$. 
\textbf{c}, Local features comprise per-nucleotide geometric and electrostatic descriptors; regional targets comprise tile-level measures of curvature, composition, duplex stability, and non-B-DNA structural propensity.}
\label{fig:overview}
\end{figure}

VANDAM complements the conventional DNA-as-a-string training paradigm with DNA molecular priors (Figure~\ref{fig:overview}). These priors provide a biophysical inductive bias without requiring an additional experimental modality.
In the following section, we first define the biological features used by VANDAM, then present Local Feature Injection and Region Property Prediction, and finally describe how these mechanisms are combined under different supervision regimes. 

\subsection{DNA Molecular Priors}
\label{sec:features}

We define a mapping from each nucleotide in a DNA sequence to a set of molecular properties. We use the 13 structural features available in DNAshapeR~\citep{chiu2016dnashaper}, whose original four-feature framework was introduced by \citet{zhou2013dnashape} and expanded to 13 features by \citet{li2017expanding}, plus minor-groove electrostatic potential (EP) from \citet{chiu2017genome}, for a total of 14 sequence-derived profiles. The features are organized into three groups:
\begin{itemize}
    \item[$\bullet$] Minor-groove environment: Minor-groove width (MGW) and electrostatic potential (EP).
    \item[$\bullet$] Intra-base-pair geometry: Propeller twist (ProT), stretch, shear, buckle, opening, and stagger.
    \item[$\bullet$] Inter-base-pair geometry: Helical twist (HelT), roll, tilt, rise, shift, and slide.
\end{itemize}
All local properties are deterministic, model-derived molecular priors rather than direct measurements and do not capture cell-specific context, environmental conditions, or conformational variability.
Appendix~\ref{app:local-features} details the complete local feature catalog.

\subsection{Local Feature Injection}
\label{sec:lfi}

Local feature injection supplies the backbone with biological priors at the input without modifying its internal blocks. Let
\(x=(x_1,\ldots,x_L)\) denote a DNA sequence of \(L\) nucleotides. Let $\phi$ be a feature map assigning each base-pair in $x$ to a vector in $\mathbb{R}^{F}$ (\S\ref{sec:features}). Because some descriptors depend on a local sequence window, \(\phi(x)_i\) may depend on the sequence context surrounding \(x_i\), rather than on \(x_i\) alone.

Let \(\operatorname{Tok}(x)=(t_1,\ldots,t_M)\) denote the tokenization of \(x\), where \(M\) need not equal \(L\). For each sequence-bearing token \(t_j\), let \(S_j \subseteq \{1,\ldots,L\}\) denote the set of nucleotide positions spanned by that token. We align the nucleotide-level features with the token sequence by mean-pooling the feature vectors over each token span:
\begin{align}
\bar{\phi}(x)_j
&=
\frac{1}{|S_j|}
\sum_{i \in S_j} \phi(x)_i,
\qquad
\bar{\phi}(x)_j \in \mathbb{R}^{F}.
\end{align}
Under single-nucleotide tokenization, each sequence-bearing token spans one nucleotide, and therefore its token-aligned biological feature vector is equal to the feature vector of that nucleotide. Features are computed at nucleotide resolution before aggregation, preserving the local sequence context that may cross token boundaries. 

We encode each token-aligned feature vector with a single linear layer:
\begin{align}
h_j
&=
W_f \, \bar{\phi}(x)_j + b_f,
\qquad
h_j, b_f \in \mathbb{R}^{H},
\quad
W_f \in \mathbb{R}^{H \times F}.
\end{align}

Let \(e(t_j) \in \mathbb{R}^{D}\) denote the backbone embedding of token
\(t_j\). We concatenate the feature embedding with the token embedding and
project the result back to the backbone's input width:
\begin{align}
m_j
&=
W_m
\left[
e(t_j) \,\|\, h_j
\right]
+b_m,
\qquad
m_j \in \mathbb{R}^{D},
W_m \in \mathbb{R}^{D \times (D+H)}.
\end{align}

The mixed embeddings \(m_j\) replace the backbone's standard embedding outputs. Biological features are computed only from nucleotides visible under the native MLM or NTP objective, preventing leakage. We first train the injector against a frozen backbone and then train both jointly; Appendix~\ref{app:local-feature-injection} gives the masking rules, warm-up procedure, token-level alignment, and backbone-specific adaptations.

\subsection{Region Property Prediction}
\label{sec:rpp}

A natural first approach to representation-level supervision is masked feature modeling (MFM), which predicts local molecular properties at masked positions.
However, each target is determined by a short sequence window, much of which remains visible after masking.
MFM therefore does not need to integrate evidence beyond that window, making any downstream gain difficult to distinguish from generic local auxiliary regularization.
Motivated by this limitation, RPP shifts supervision from local features to tile-level properties defined over genomic regions, encouraging the pooled representation to retain information distributed across a broader sequence context.

\subsubsection{Regional Properties}
For each genomic region, we compute six sequence-derived properties: intrinsic curvature, AT content, melting propensity, G-quadruplex propensity, cruciform propensity, and Z-DNA propensity. 
Together, these targets span regional geometry, composition, duplex stability, and non-B-DNA structural potential which quantifies sequence-based propensity. Appendix~\ref{app:regional-features} gives the exact definition, unit, 
and biological motivation for each selected property.

\subsubsection{Learning with Regional Properties}
RPP  directly rewards pooled representations for retaining  molecular properties, rather than leaving their preservation to emerge from the native token objective.
Given a DNA sequence, we partition it into fixed non-overlapping tiles $\mathcal{T}$ (regions).
For each tile, the backbone's hidden states are pooled to a single vector, $z_j = \mathrm{pool}\big(\{H_i : i \in \mathcal{T}_j\}\big)$, where $\mathcal{T}_j$ contains the token indices of tile $j$ and $H_i$ is the backbone hidden state at position $i$.

A lightweight linear head predicts a vector of properties derived from the true sequence, $\hat{y}_j = W_{\mathrm{RPP}} z_j + b_{\mathrm{RPP}}$, where $W_{\mathrm{RPP}} \in \mathbb{R}^{N \times D}$, $b_{\mathrm{RPP}} \in \mathbb{R}^{N}$, and $N$ is the number of features used. Let $\mathcal{V}$ be the set of valid tiles in a minibatch; the RPP loss is the mean squared error, averaged over valid tiles and target dimensions:
\begin{equation}
    \mathcal{L}_{\mathrm{RPP}}
    = \frac{1}{N}\sum_{r=1}^{N}
      \frac{1}{|\mathcal{V}|}\sum_{j\in\mathcal{V}}
      \left(\hat{y}_{j,r}-y_{j,r}\right)^2.
\end{equation}

The RPP loss is combined with the native token loss, $\mathcal{L}_{\mathrm{token}}$  (typically, MLM or NTP), using a linearly warmed coefficient,
\begin{equation}
    \mathcal{L} = \mathcal{L}_{\mathrm{token}}
    + \lambda_{\mathrm{RPP}}(s)\mathcal{L}_{\mathrm{RPP}},
    \qquad
    \lambda_{\mathrm{RPP}}(s)
    = \lambda_{\max}\min\!\left(1,\frac{s}{1000}\right)
\end{equation}
where $s$ is the optimizer step. Thus, $\lambda_{\mathrm{RPP}}$ increases linearly from zero during the first 1{,}000 steps and remains at its configured value thereafter.
If $\mathcal{V}=\emptyset$, the RPP loss is omitted for that minibatch and optimization proceeds using only $\mathcal{L}_{\mathrm{token}}$. Both terms read from the same forward pass, so RPP adds no 
additional encoder evaluation. Appendix~\ref{app:rpp-implementation} gives target transformation and calibration details, precise tiling and validity rules, and backbone-specific BPE alignment details.

\subsection{VANDAM Training Regimes}
\label{sec:vandam-regimes}

The appropriate mechanism for incorporating molecular priors depends on whether the training objective makes the supplied information consequential.
We therefore consider two configurations. In the self-supervised pre-training phase, \emph{VANDAM-PT} applies RPP without LFI. 
We omit LFI because its local descriptors are deterministic functions of context available to the native token objective and are not themselves supervised.
In multi-task tuning, when functional task labels are available alongside the native token objective, \emph{VANDAM-MT} combines LFI with RPP.
Task gradients can reward use of the injected descriptors for functional prediction, while RPP directly supervises retention of regional properties.

%%%%%%%%%%%%%%%%%%%%%%%%%%%%%%%%%%%%%%%%%%%%%%%%%%%%%%%%%%%%
\section{Experiments}
%%%%%%%%%%%%%%%%%%%%%%%%%%%%%%%%%%%%%%%%%%%%%%%%%%%%%%%%%%%%

\subsection{Setup}
\label{sec:setup}

\paragraph{Models.}
We evaluate VANDAM on four backbones spanning four distinct architecture families to assess whether the effect generalizes beyond a single architectural choice: NTv3-8M \citep{boshar2025foundational}, a U-Net convolutional--transformer hybrid; Caduceus-PS \citep{schiff2024caduceus}, a bi-directional Mamba state-space model; DNABERT-2-117M \citep{zhou2024dnabert}, a flat transformer; and HyenaDNA-small-32k \citep{nguyen2023hyenadna}, a long-range Hyena-operator model. We use the shorter family names Caduceus and HyenaDNA thereafter.

\begin{table}[h]
\centering
\caption{Backbone models used in all experiments. Parameter counts are trainable backbone parameters in our implementation, excluding VANDAM and downstream heads.}
\label{tab:models}
\small
\begin{tabular}{lllll}
\toprule
\textbf{Model} & \textbf{Params} & \textbf{Architecture} & \textbf{Tokenizer} & \textbf{Pre-training objective} \\
\midrule
NTv3-8M        & 8M    & U-Net conv-transformer & Single-nucleotide & MLM \\
Caduceus-PS       & 7.73M & Bi-directional Mamba   & Single-nucleotide & MLM \\
DNABERT-2         & 117M  & Flat transformer       & BPE               & MLM \\
HyenaDNA  & 3.28M & Hyena operator         & Single-nucleotide & NTP \\
\bottomrule
\end{tabular}
\end{table}

\paragraph{Pre-training.}
We apply continual pre-training to all models on OpenGenome2 \citep{brixi2026genome}, a sequence-only multi-species corpus of roughly 8.8 trillion nucleotides spanning bacteria, archaea, eukarya and phage.
We compare VANDAM-PT with its released checkpoint and Baseline-PT, its compute-matched control using only the native token objective.
Matching the data and compute budget isolates the effect of the injected biology from the effect of simply training longer. Appendix~\ref{app:training-details} gives the full training configuration.

\paragraph{Multi-task training.}
As a second regime, we perform mixed-objective multi-task continual training using eight contamination-controlled functional tasks disjoint from 
the evaluation suite (Appendix~\ref{app:pool}). A task-batch fraction $\rho_{\mathrm{task}}$ controls the proportion of optimization steps assigned 
to labeled task batches; the remaining updates retain the backbone's native sequence objective. Both Baseline-MT and VANDAM-MT use the same eight 
task-specific supervised heads, one for each task in the multi-task pool; VANDAM-MT additionally includes the shared RPP head. VANDAM-MT applies 
LFI to both batch types and RPP only to sequence-only batches, where it augments the native MLM or NTP loss. 
Baseline-MT is its compute-matched control, using the same data, schedule, and task mixture without LFI or RPP. Appendix~\ref{app:training-details} gives the schedule, head warm-up, and compute budget.

\paragraph{Evaluation and statistical analysis.}
We evaluate through GFMBench-API \citep{larey2026gfmbench} on nine held-out tasks: four zero-shot variant-effect tasks, 
scored by cosine dissimilarity between pooled reference and alternate-sequence embeddings, and five tasks using frozen representations 
with linear probes. We use AUROC for variant-effect and classification tasks, and macro Pearson correlation for CAGE expression. 
Appendix~\ref{app:eval-tasks} gives the task sources, sample sizes, metrics, and probe protocol.

For each task--backbone comparison, we evaluate the paired score difference $\Delta=\mathrm{VANDAM}-\mathrm{Baseline}$; 
for linear-probe tasks, this is averaged across five matched probe seeds. We construct 95\% confidence intervals from 10,000 paired bootstrap 
replicates, resampling test units for all tasks and the five probe seeds for linear-probe tasks. Each replicate uses the same sampled test 
units for VANDAM and Baseline. Zero-shot tasks therefore resample test units only.
We obtain one-sided $p$-values from 10,000 paired randomization replicates that swap VANDAM and Baseline predictions within each test unit; 
for linear-probe tasks, each swap is shared across seeds. We apply Benjamini--Hochberg (BH) adjustment separately within each training regime and direction. 
A VANDAM advantage requires a positive mean paired difference, a 95\% confidence interval wholly above zero, and an adjusted positive-direction 
$q<0.05$; we use the mirrored criterion for Baseline advantages.
Appendix~\ref{app:paired-inference} provides the complete inferential procedure and comparison-level results.

%%%%%%%%%%%%%%%%%%%%%%%%%%%%%%%%%%%%%%%%%%%%%%%%%%%%%%%%%%%%
\subsection{Downstream Performance and Component Ablations}

Tables~\ref{tab:pretrain-pertask} and~\ref{tab:multitask} report the per-task comparisons for continual pre-training (Baseline-PT versus VANDAM-PT) and multi-task training (Baseline-MT versus VANDAM-MT), respectively.
Scores for frozen linear-probe tasks are means across the five matched probe seeds.
To probe the role of target locality, Appendix~\ref{app:mfm} contrasts RPP with MFM and a random features control, 
showing how downstream gains change when supervision moves from local features to regional properties. 
To understand the contribution of each component under the two training regimes, Table~\ref{tab:pretrain-ablation} compares LFI and 
RPP in isolation and in combination on NTv3-8M.
Appendix~\ref{app:feature-pathway} further characterizes the learned LFI pathway, while Appendix~\ref{app:rpp-target-ablation} separately ablates the six RPP targets.

\begin{table}[t]
\centering
\caption{Pre-training results per task. Bold denotes the higher central score; superscript $+$/$-$ denotes a BH-adjusted paired difference favoring VANDAM/Baseline.}
\label{tab:pretrain-pertask}
\scriptsize
\setlength{\tabcolsep}{2pt}
\begin{tabular}{lcccccccc}
\toprule
& \multicolumn{2}{c}{\textbf{NTv3-8M}} & \multicolumn{2}{c}{\textbf{Caduceus}} & \multicolumn{2}{c}{\textbf{DNABERT-2}} & \multicolumn{2}{c}{\textbf{HyenaDNA}} \\
\cmidrule(lr){2-3} \cmidrule(lr){4-5} \cmidrule(lr){6-7} \cmidrule(lr){8-9}
\textbf{Task} & \shortstack{Baseline-\\PT} & \shortstack{VANDAM-\\PT} & \shortstack{Baseline-\\PT} & \shortstack{VANDAM-\\PT} & \shortstack{Baseline-\\PT} & \shortstack{VANDAM-\\PT} & \shortstack{Baseline-\\PT} & \shortstack{VANDAM-\\PT} \\
\midrule
Clinical pathogenicity (ClinVar) & 58.0 & \textbf{61.7}$^\dagger$ & 60.5 & \textbf{61.9}$^\dagger$ & 59.3 & \textbf{65.9}$^\dagger$ & 66.9 & \textbf{75.3}$^\dagger$ \\
Disease variants (OMIM)      & 49.8 & \textbf{59.0}$^\dagger$ & 48.6 & \textbf{55.0}$^\dagger$ & \textbf{55.7}$^\ddagger$ & 51.2 & \textbf{52.2} & 52.1 \\
Mendelian variants (TraitGym) & 49.7 & \textbf{55.8}$^\dagger$ & 47.2 & \textbf{55.9}$^\dagger$ & \textbf{54.2} & 53.7 & 53.9 & \textbf{54.2} \\
BRCA1 variant effect         & 50.9 & \textbf{51.6} & 49.6 & \textbf{54.0}$^\dagger$ & 56.9 & \textbf{57.6} & 51.9 & \textbf{55.3}$^\dagger$ \\
Transcription-factor binding & 79.7 & \textbf{80.8}$^\dagger$ & 68.9 & \textbf{75.2}$^\dagger$ & 79.0 & \textbf{80.3}$^\dagger$ & 73.7 & \textbf{80.0}$^\dagger$ \\
Regulatory variants (eQTL)   & 68.3 & \textbf{70.2}$^\dagger$ & \textbf{67.0} & 66.5 & 69.1 & \textbf{71.9}$^\dagger$ & 68.2 & \textbf{69.3}$^\dagger$ \\
Gene-expression variants     & 61.8 & \textbf{63.3}$^\dagger$ & \textbf{62.4} & 61.8 & 65.2 & \textbf{66.3}$^\dagger$ & 63.3 & \textbf{63.7} \\
Chromatin state (histone)    & 66.0 & \textbf{68.1}$^\dagger$ & 60.2 & \textbf{65.7}$^\dagger$ & 66.1 & \textbf{69.8}$^\dagger$ & \textbf{66.2}$^\ddagger$ & 63.4 \\
CAGE expression                   & 24.2 & \textbf{34.9}$^\dagger$ & \textbf{23.4}$^\ddagger$ & 22.2 & 34.1 & \textbf{34.5}$^\dagger$ & 23.5 & \textbf{30.5}$^\dagger$ \\
\bottomrule
\end{tabular}
\end{table}

%%%%%%%%%%%%%%%%%%%%%%%%%%%%%%%%%%%%%%%%%%%%%%%%%%%%%%%%%%%%
The evaluation suite covers clinical and Mendelian variant pathogenicity, experimentally measured BRCA1 effects, transcription-factor binding, regulatory and gene-expression variant effects, chromatin-state prediction, and CAGE expression.
VANDAM has positive point-estimate improvements in 29 of 36 pre-training and 33 of 36 multi-task task--backbone comparisons, spanning all four backbones.
After paired inference and within-regime BH adjustment, VANDAM shows statistically significant improvements on 25 pre-training and 23 multi-task comparisons. In contrast, the compute-matched baseline is significantly superior on three pre-training and one multi-task training comparison.
This trend holds across convolutional, state-space, transformer, and Hyena architectures; single-nucleotide and BPE tokenization; and both masked-language and next-token prediction objectives.
Together, the results support molecular grounding as a broadly useful auxiliary signal, both alone and when coupled with functional supervision.
Intrinsic token-objective metrics (e.g. perplexity) vary by backbone and regime rather than increasing uniformly (Appendix~\ref{app:intrinsic}), indicating that the downstream gains are not explained solely by improved token-identity prediction.

\begin{table}[t]
\centering
\caption{Multi-task training results per task. Bold denotes the higher central score; superscript $+$/$-$ denotes a BH-adjusted paired difference favoring VANDAM/Baseline.}
\label{tab:multitask}
\scriptsize
\setlength{\tabcolsep}{2pt}
\begin{tabular}{lcccccccc}
\toprule
& \multicolumn{2}{c}{\textbf{NTv3-8M}} & \multicolumn{2}{c}{\textbf{Caduceus}} & \multicolumn{2}{c}{\textbf{DNABERT-2}} & \multicolumn{2}{c}{\textbf{HyenaDNA}} \\
\cmidrule(lr){2-3} \cmidrule(lr){4-5} \cmidrule(lr){6-7} \cmidrule(lr){8-9}
\textbf{Task} & \shortstack{Baseline-\\MT} & \shortstack{VANDAM-\\MT} & \shortstack{Baseline-\\MT} & \shortstack{VANDAM-\\MT} & \shortstack{Baseline-\\MT} & \shortstack{VANDAM-\\MT} & \shortstack{Baseline-\\MT} & \shortstack{VANDAM-\\MT} \\
\midrule
Clinical pathogenicity (ClinVar) & 68.1 & \textbf{70.2}$^\dagger$ & \textbf{54.4}$^\ddagger$ & 51.4 & 59.5 & \textbf{69.7}$^\dagger$ & 55.7 & \textbf{71.2}$^\dagger$ \\
Disease variants (OMIM)      & 54.9 & \textbf{57.4} & 47.4 & \textbf{53.1}$^\dagger$ & 52.0 & \textbf{66.4}$^\dagger$ & 47.9 & \textbf{51.0} \\
Mendelian variants (TraitGym) & 51.5 & \textbf{56.8}$^\dagger$ & 47.8 & \textbf{55.6}$^\dagger$ & 55.0 & \textbf{57.5} & 51.4 & \textbf{54.8} \\
BRCA1 variant effect         & 51.9 & \textbf{54.6} & 51.7 & \textbf{52.5} & 52.5 & \textbf{56.2}$^\dagger$ & 51.6 & \textbf{53.4} \\
Transcription-factor binding & 78.7 & \textbf{79.8}$^\dagger$ & 53.7 & \textbf{66.4}$^\dagger$ & 78.1 & \textbf{82.6}$^\dagger$ & 75.2 & \textbf{81.0}$^\dagger$ \\
Regulatory variants (eQTL)   & 68.1 & \textbf{69.3}$^\dagger$ & 60.9 & \textbf{63.0}$^\dagger$ & 68.8 & \textbf{71.8}$^\dagger$ & 69.7 & \textbf{70.0} \\
Gene-expression variants     & 62.6 & \textbf{63.1} & 57.8 & \textbf{59.9}$^\dagger$ & 68.2 & \textbf{69.4}$^\dagger$ & \textbf{65.8} & 64.9 \\
Chromatin state (histone)    & 68.3 & \textbf{69.3}$^\dagger$ & 52.9 & \textbf{55.4}$^\dagger$ & 67.2 & \textbf{72.6}$^\dagger$ & 68.2 & \textbf{69.2}$^\dagger$ \\
CAGE expression                   & 31.7 & \textbf{35.9}$^\dagger$ & 13.7 & \textbf{14.6} & \textbf{36.9} & 36.6 & 20.5 & \textbf{30.8}$^\dagger$ \\
\bottomrule
\end{tabular}
\end{table}

\begin{table}[t]
\centering
\caption{Component ablation across pre-training and multi-task training on NTv3-8M. Values are mean AUROC over the eight AUROC tasks.}
\label{tab:pretrain-ablation}
\scriptsize
\setlength{\tabcolsep}{2pt}
\begin{tabular}{lcc}
\toprule
\textbf{Configuration} & \textbf{Pre-training} & \textbf{Multi-task} \\
\midrule
Original checkpoint              & 59.5 & 59.5 \\
Baseline-PT / Baseline-MT        & 60.5 & 63.0 \\
\midrule
LFI only                         & 60.5 & 64.2 \\
RPP only                         & \textbf{63.8} & 63.2 \\
LFI + RPP                        & 63.4 & \textbf{65.1} \\
\midrule
RPP without MLM loss, masking retained & 60.2 & 60.2 \\
RPP without MLM loss, no masking       & 59.7 & 60.1 \\
\bottomrule
\end{tabular}
\end{table}

\subsection{Does Biological Content Matter?}
\label{sec:ablation-content}

We test whether VANDAM's gains arise from biological content rather than added capacity or optimization pathways. 
For LFI, we replace $\phi(x)_i$ with a random vector of the same dimension. 
For RPP, we use matched shuffled-table and algorithmic regional targets that remove the biological sequence-to-property mapping while preserving 
the regional prediction problem; Appendix~\ref{app:biological-controls} gives the complete control construction.

\begin{figure}[H]
\centering
\includegraphics[width=0.93\linewidth]{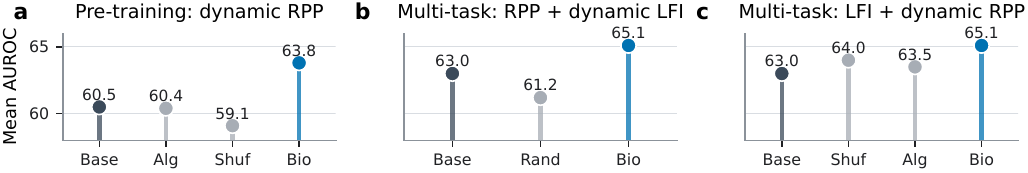}
\vspace{-0.4em}
\caption{Biological-content controls on NTv3-8M (mean AUROC over eight tasks). \textbf{a}, Pre-training RPP-target controls. \textbf{b}, Multi-task LFI controls with biological RPP. \textbf{c}, Multi-task RPP-target controls with biological LFI.}
\label{fig:biological-content-controls}
\end{figure}

Figure~\ref{fig:biological-content-controls}a shows that, during pre-training, biological RPP exceeds the algorithmic and shuffled-table controls by 3.4 
and 4.7 mean-AUROC points, respectively; thus, an auxiliary regional-prediction task is insufficient without biologically meaningful targets. 
Under multi-task training, randomizing LFI reduces mean AUROC by 3.9 points (Figure~\ref{fig:biological-content-controls}b), 
while replacing biological RPP with shuffled-table or algorithmic targets reduces it by 1.1 and 1.6 points, respectively 
(Figure~\ref{fig:biological-content-controls}c).

%%%%%%%%%%%%%%%%%%%%%%%%%%%%%%%%%%%%%%%%%%%%%%%%%%%%%%%%%%%%
\subsection{RPP Transfers to Unseen Molecular Descriptors}
\label{sec:qdna-transfer}

The downstream evaluations above establish that VANDAM improves task performance, but do not show whether the use of molecular priors generalizes to other unseen molecular properties. To examine what changed in the representation, we turn to qDNA, which models sequence-dependent charge dynamics to compute model-derived electronic and photophysical observables of hole and electron localization, transfer, exciton decay, and electron--hole separation \citep{herb2025quantumdna,hawke2010electronic}.
We use these quantities as held-out test-time molecular descriptors, which provide a physically distinct view of DNA from the properties used in RPP.

We fit ridge regression models using representations from Baseline-PT and VANDAM-PT to predict each one of the eight qDNA descriptors averaged over 300-bp tiles. 
Since qDNA descriptors may be linearly correlated with the six properties used as RPP targets, we further fit an OLS model on those properties to predict qDNA features. We then subtract its prediction from the ground-truth qDNA values and use the resulting residual as a new target for ridge probing. This ensures that the reported improvement (Table~\ref{tab:qdna-transfer}) is only attributed to the learned representations rather than to linear correlation between the two set of properties. Appendix~\ref{app:qdna-probes} gives the complete target construction, probing, and inference protocol.

\begin{table}[!ht]
\centering
\caption{qDNA transfer gain ($\Delta R^2$, VANDAM-PT minus Baseline-PT). Bold denotes statistical significant improvement.}
\label{tab:qdna-transfer}
\small
\setlength{\tabcolsep}{4pt}
\renewcommand{\arraystretch}{0.92}
\begin{tabular}{@{}lcclcc@{}}
\toprule
\textbf{Descriptor} & \textbf{Original} & \textbf{Residual} & \textbf{Descriptor} & \textbf{Original} & \textbf{Residual} \\
\midrule
Hole central population   & \textbf{0.060} & \textbf{0.054} & Electron central population   & \textbf{0.083} & \textbf{0.057} \\
Hole maximum transfer     & $-0.041$       & $-0.032$       & Electron maximum transfer     & $-0.055$       & $-0.046$ \\
Hole delocalization (IPR) & \textbf{0.071} & \textbf{0.091} & Electron delocalization (IPR) & \textbf{0.067} & \textbf{0.051} \\
Exciton lifetime          & \textbf{0.073} & \textbf{0.084} & Electron--hole separation     & \textbf{0.069} & \textbf{0.076} \\
\bottomrule
\end{tabular}
\vspace{-0.5em}
\end{table}

VANDAM-PT improves the predictability for six out of eight qDNA descriptors, both for original and residualized. A gain is considered statistically significant and reported as an improvement when the BH-adjusted $q\leq0.05$ and $\Delta R^2\geq0.02$. Full results and 95\% CIs are in Appendix Table~\ref{tab:qdna-probes}.

These results show that VANDAM improves linear access to electronic and photophysical descriptors that were never used for training. Their persistence after residualization indicates that the transfer is not explained solely by a linear recombination of the six supervised RPP targets. Thus, within NTv3-8M, VANDAM generalizes beyond its training properties to unseen molecular descriptors.

%%%%%%%%%%%%%%%%%%%%%%%%%%%%%%%%%%%%%%%%%%%%%%%%%%%%%%%%%%%%
\section{Discussion}
%%%%%%%%%%%%%%%%%%%%%%%%%%%%%%%%%%%%%%%%%%%%%%%%%%%%%%%%%%%%

Across four architecture families, including two tokenization schemes, and both MLM and NTP objectives, molecular grounding improves performance in most downstream task comparisons. This consistent performance supports an objective--task alignment interpretation: token recovery is a useful general-purpose pre-training signal, but it does not directly require representations to preserve molecular properties relevant to downstream functional and phenotypic prediction. The pre-training RPP-without-MLM ablation underperforms Baseline-PT (Table~\ref{tab:pretrain-ablation}), further indicating that molecular grounding complements rather than replaces the token objective.

This complementarity is best understood through the role of the training objective. VANDAM does not introduce information unavailable from sequence: every local descriptor and regional property used in this work is a deterministic function of the input. The operative variable is instead what the objective obliges the representation to preserve. Making a property available is passive and can be ignored; attaching a loss to it, or attaching labels whose prediction it eases, makes its use consequential. This differs from settings in which a supplied structural prior contains information not directly present in the model's sequence input, whether that information is experimentally measured or obtained from another source. In the NTv3-8M component ablation, LFI alone matches Baseline-PT during continual pre-training (both $60.5$ mean AUROC), but improves over Baseline-MT during multi-task training ($64.2$ versus $63.0$; Table~\ref{tab:pretrain-ablation}). This pattern is consistent with a simple asymmetry: the native token objective can be optimized from the same bounded local context from which an injected descriptor is computed, without requiring the representation to retain that descriptor. Functional labels, by contrast, can reward using a biologically meaningful featurization to predict targets not fixed by the local window.

Together, the biological-content controls and qDNA probing suggest that RPP does more than make its six supervised targets linearly accessible: it promotes representations that retain broader molecular structure. The persistence of transfer after residualization strengthens this interpretation.

\textbf{Limitations and future work.} The present study focuses on molecular properties defined over local genomic tiles and on comparatively small backbones with sequence contexts of at most 8{,}192\,bp, enabling controlled compute-matched comparisons across architectures. We therefore do not establish how VANDAM behaves as the spatial scale of its molecular priors, model capacity, and sequence context grow together. Under MLM, zeroing descriptors whose pentamer intersects a mask makes approximately 47\% of interior LFI vectors zero. Appendix~\ref{app:lfi-mask-sensitivity} evaluates feature-only imputation on NTv3-8M; the effects are task-specific and were not tested across backbones. Extending these masking-policy experiments alongside supervision for long-range and three-dimensional genome organization is an important direction for future work.

%%%%%%%%%%%%%%%%%%%%%%%%%%%%%%%%%%%%%%%%%%%%%%%%%%%%%%%%%%%%
\section{Conclusion}
%%%%%%%%%%%%%%%%%%%%%%%%%%%%%%%%%%%%%%%%%%%%%%%%%%%%%%%%%%%%

VANDAM reframes genomic foundation-model training around DNA as a molecule as well as a sequence.
By directly supervising regional molecular properties during self-supervised continual pre-training, and by incorporating local molecular descriptors when functional labels can reward their use, it provides objective-appropriate routes for grounding representations in sequence-derived biophysics.
Across diverse architectures, tokenizers, and token-identity objectives, these interventions improve downstream genomic representations; RPP also increases access to held-out molecular descriptors.
More broadly, molecular priors can help genomic foundation models capture sequence-dependent physical properties of DNA.

\newpage
\bibliographystyle{iclr2027_conference}
\bibliography{refs}

\begin{thebibliography}{40}
\providecommand{\natexlab}[1]{#1}
\providecommand{\url}[1]{\texttt{#1}}
\expandafter\ifx\csname urlstyle\endcsname\relax
  \providecommand{\doi}[1]{doi: #1}\else
  \providecommand{\doi}{doi: \begingroup \urlstyle{rm}\Url}\fi

\bibitem[Albors et~al.(2025)Albors, Li, Benegas, Ye, and
  Song]{albors2025phylogenetic}
C~Albors, JC~Li, G~Benegas, C~Ye, and YS~Song.
\newblock A phylogenetic approach to genomic language modeling in research in
  computational molecular biology, 2025.

\bibitem[Bedrat et~al.(2016)Bedrat, Lacroix, and Mergny]{bedrat2016re}
Amina Bedrat, Laurent Lacroix, and Jean-Louis Mergny.
\newblock Re-evaluation of g-quadruplex propensity with g4hunter.
\newblock \emph{Nucleic acids research}, 44\penalty0 (4):\penalty0 1746--1759,
  2016.

\bibitem[Benegas et~al.(2025{\natexlab{a}})Benegas, Albors, Aw, Ye, and
  Song]{benegas2025dna}
Gonzalo Benegas, Carlos Albors, Alan~J Aw, Chengzhong Ye, and Yun~S Song.
\newblock A dna language model based on multispecies alignment predicts the
  effects of genome-wide variants.
\newblock \emph{Nature Biotechnology}, 43\penalty0 (12):\penalty0 1960--1965,
  2025{\natexlab{a}}.

\bibitem[Benegas et~al.(2025{\natexlab{b}})Benegas, Eraslan, and
  Song]{benegas2025benchmarking}
Gonzalo Benegas, G{\"o}kcen Eraslan, and Yun~S Song.
\newblock Benchmarking dna sequence models for causal regulatory variant
  prediction in human genetics.
\newblock \emph{bioRxiv}, 2025{\natexlab{b}}.

\bibitem[Benegas et~al.(2025{\natexlab{c}})Benegas, Ye, Albors, Li, and
  Song]{benegas2025genomic}
Gonzalo Benegas, Chengzhong Ye, Carlos Albors, Jianan~Canal Li, and Yun~S Song.
\newblock Genomic language models: opportunities and challenges.
\newblock \emph{Trends in Genetics}, 41\penalty0 (4):\penalty0 286--302,
  2025{\natexlab{c}}.

\bibitem[Boshar et~al.(2025)Boshar, Evans, Tang, Picard, Adel, Lorbeer, Rajesh,
  Karch, Sidbon, Emms, et~al.]{boshar2025foundational}
Sam Boshar, Benjamin Evans, Ziqi Tang, Armand Picard, Yanis Adel, Franziska~K
  Lorbeer, Chandana Rajesh, Tristan Karch, Shawn Sidbon, David Emms, et~al.
\newblock A foundational model for joint sequence-function multi-species
  modeling at scale for long-range genomic prediction.
\newblock \emph{BioRxiv}, pp.\  2025--12, 2025.

\bibitem[Br{\'a}zda et~al.(2011)Br{\'a}zda, Laister, Jagelsk{\'a}, and
  Arrowsmith]{brazda2011cruciform}
V{\'a}clav Br{\'a}zda, Rob~C Laister, Eva~B Jagelsk{\'a}, and Cheryl
  Arrowsmith.
\newblock Cruciform structures are a common dna feature important for
  regulating biological processes.
\newblock \emph{BMC molecular biology}, 12\penalty0 (1):\penalty0 33, 2011.

\bibitem[Brixi et~al.(2026)Brixi, Durrant, Ku, Naghipourfar, Poli, Sun,
  Brockman, Chang, Fanton, Gonzalez, et~al.]{brixi2026genome}
Garyk Brixi, Matthew~G Durrant, Jerome Ku, Mohsen Naghipourfar, Michael Poli,
  Gwanggyu Sun, Greg Brockman, Daniel Chang, Alison Fanton, Gabriel~A Gonzalez,
  et~al.
\newblock Genome modelling and design across all domains of life with evo 2.
\newblock \emph{Nature}, 652\penalty0 (8112):\penalty0 1349--1361, 2026.

\bibitem[Chiu et~al.(2016)Chiu, Comoglio, Zhou, Yang, Paro, and
  Rohs]{chiu2016dnashaper}
Tsu-Pei Chiu, Federico Comoglio, Tianyin Zhou, Lin Yang, Renato Paro, and Remo
  Rohs.
\newblock Dnashaper: an r/bioconductor package for dna shape prediction and
  feature encoding.
\newblock \emph{Bioinformatics}, 32\penalty0 (8):\penalty0 1211--1213, 2016.

\bibitem[Chiu et~al.(2017)Chiu, Rao, Mann, Honig, and Rohs]{chiu2017genome}
Tsu-Pei Chiu, Satyanarayan Rao, Richard~S Mann, Barry Honig, and Remo Rohs.
\newblock Genome-wide prediction of minor-groove electrostatic potential
  enables biophysical modeling of protein--dna binding.
\newblock \emph{Nucleic acids research}, 45\penalty0 (21):\penalty0
  12565--12576, 2017.

\bibitem[Chu et~al.(2024)Chu, Yu, Li, Huang, Shen, Cong, Zhang, and
  Wang]{chu20245}
Yanyi Chu, Dan Yu, Yupeng Li, Kaixuan Huang, Yue Shen, Le~Cong, Jason Zhang,
  and Mengdi Wang.
\newblock A 5' utr language model for decoding untranslated regions of mrna and
  function predictions.
\newblock \emph{Nature Machine Intelligence}, 6\penalty0 (4):\penalty0
  449--460, 2024.

\bibitem[Findlay et~al.(2018)Findlay, Daza, Martin, Zhang, Leith, Gasperini,
  Janizek, Huang, Starita, and Shendure]{findlay2018accurate}
Gregory~M Findlay, Riza~M Daza, Beth Martin, Melissa~D Zhang, Anh~P Leith,
  Molly Gasperini, Joseph~D Janizek, Xingfan Huang, Lea~M Starita, and Jay
  Shendure.
\newblock Accurate classification of brca1 variants with saturation genome
  editing.
\newblock \emph{Nature}, 562\penalty0 (7726):\penalty0 217--222, 2018.

\bibitem[Hawke et~al.(2010)Hawke, Kalosakas, and
  Simserides]{hawke2010electronic}
Laurence George~Demosthenis Hawke, George Kalosakas, and Constantinos
  Simserides.
\newblock Electronic parameters for charge transfer along dna.
\newblock \emph{The European Physical Journal E}, 32\penalty0 (3):\penalty0
  291--305, 2010.

\bibitem[Herb et~al.(2025)Herb, Trenti, Mantela, Simserides, Ankerhold, and
  Rossini]{herb2025quantumdna}
Dennis Herb, Marco Trenti, Marilena Mantela, Constantinos Simserides, Joachim
  Ankerhold, and Mirko Rossini.
\newblock Quantumdna: A python package for analyzing quantum charge dynamics in
  dna and exploring its biological relevance.
\newblock \emph{Computer Physics Communications}, 313:\penalty0 109626, 2025.

\bibitem[Ji et~al.(2021)Ji, Zhou, Liu, and Davuluri]{ji2021dnabert}
Yanrong Ji, Zhihan Zhou, Han Liu, and Ramana~V Davuluri.
\newblock Dnabert: pre-trained bidirectional encoder representations from
  transformers model for dna-language in genome.
\newblock \emph{Bioinformatics}, 37\penalty0 (15):\penalty0 2112--2120, 2021.

\bibitem[Kelley et~al.(2018)Kelley, Reshef, Bileschi, Belanger, McLean, and
  Snoek]{kelley2018sequential}
David~R Kelley, Yakir~A Reshef, Maxwell Bileschi, David Belanger, Cory~Y
  McLean, and Jasper Snoek.
\newblock Sequential regulatory activity prediction across chromosomes with
  convolutional neural networks.
\newblock \emph{Genome research}, 28\penalty0 (5):\penalty0 739--750, 2018.

\bibitem[Larey et~al.(2026{\natexlab{a}})Larey, Dahan, Bleiweiss, Kellerman,
  Leib, Nayshool, Ofer, Zinger, Dominissini, Rechavi,
  et~al.]{larey2026gfmbench}
Ariel Larey, Elay Dahan, Amit Bleiweiss, Raizy Kellerman, Guy Leib, Omri
  Nayshool, Dan Ofer, Tal Zinger, Dan Dominissini, Gideon Rechavi, et~al.
\newblock Gfmbench-api: A standardized interface for benchmarking genomic
  foundation models.
\newblock \emph{Biorxiv}, pp.\  2026--02, 2026{\natexlab{a}}.

\bibitem[Larey et~al.(2026{\natexlab{b}})Larey, Dahan, Bleiweiss, Kellerman,
  Leib, Nayshool, Ofer, Zinger, Dominissini, Rechavi, et~al.]{larey2026jepa}
Ariel Larey, Elay Dahan, Amit Bleiweiss, Raizy Kellerman, Guy Leib, Omri
  Nayshool, Dan Ofer, Tal Zinger, Dan Dominissini, Gideon Rechavi, et~al.
\newblock Jepa-dna: Grounding genomic foundation models through joint-embedding
  predictive architectures.
\newblock \emph{arXiv preprint arXiv:2602.17162}, 2026{\natexlab{b}}.

\bibitem[Li et~al.(2017)Li, Sagendorf, Chiu, Pasi, Perez, and
  Rohs]{li2017expanding}
Jinsen Li, Jared~M Sagendorf, Tsu-Pei Chiu, Marco Pasi, Alberto Perez, and Remo
  Rohs.
\newblock Expanding the repertoire of dna shape features for genome-scale
  studies of transcription factor binding.
\newblock \emph{Nucleic acids research}, 45\penalty0 (22):\penalty0
  12877--12887, 2017.

\bibitem[Li et~al.(2024)Li, Chiu, and Rohs]{li2024predicting}
Jinsen Li, Tsu-Pei Chiu, and Remo Rohs.
\newblock Predicting dna structure using a deep learning method.
\newblock \emph{Nature communications}, 15\penalty0 (1):\penalty0 1243, 2024.

\bibitem[Liu et~al.(2026)Liu, Zhang, Lin, Pinello, Ying, and Zhao]{liu2026pre}
Tianyu Liu, Xiangyu Zhang, Jiecong Lin, Luca Pinello, Rex Ying, and Hongyu
  Zhao.
\newblock Pre-training genomic language model with variants for better modeling
  functional genomics.
\newblock \emph{npj Artificial Intelligence}, 2\penalty0 (1):\penalty0 46,
  2026.

\bibitem[Long et~al.(2025)Long, Su, Xiong, and Zhang]{long2025mutbert}
Weicai Long, Houcheng Su, Jiaqi Xiong, and Yanlin Zhang.
\newblock Mutbert: probabilistic genome representation improves genomics
  foundation models.
\newblock \emph{Bioinformatics}, 41\penalty0 (Supplement\_1):\penalty0
  i294--i303, 2025.

\bibitem[Lu et~al.(2025)Lu, Liu, Lin, and Brandes]{lu2025genomic}
Baiyu Lu, Xueshen Liu, Po-Yu Lin, and Nadav Brandes.
\newblock Genomic heterogeneity inflates the performance of variant
  pathogenicity predictions.
\newblock \emph{bioRxiv}, pp.\  2025--09, 2025.

\bibitem[Mathelier et~al.(2016)Mathelier, Xin, Chiu, Yang, Rohs, and
  Wasserman]{mathelier2016dna}
Anthony Mathelier, Beibei Xin, Tsu-Pei Chiu, Lin Yang, Remo Rohs, and Wyeth~W
  Wasserman.
\newblock Dna shape features improve transcription factor binding site
  predictions in vivo.
\newblock \emph{Cell systems}, 3\penalty0 (3):\penalty0 278--286, 2016.

\bibitem[Medvedev et~al.(2025)Medvedev, Viswanathan, Kanithi, Vishniakov,
  Munjal, Christophe, Pimentel, Rajan, and Khan]{medvedev2025biotoken}
Aleksandr Medvedev, Karthik Viswanathan, Praveenkumar Kanithi, Kirill
  Vishniakov, Prateek Munjal, Clement Christophe, Marco~AF Pimentel, Ronnie
  Rajan, and Shadab Khan.
\newblock Biotoken and biofm--biologically-informed tokenization enables
  accurate and efficient genomic foundation models.
\newblock \emph{bioRxiv}, pp.\  2025--03, 2025.

\bibitem[Nguyen et~al.(2023)Nguyen, Poli, Faizi, Thomas, Wornow, Birch-Sykes,
  Massaroli, Patel, Rabideau, Bengio, et~al.]{nguyen2023hyenadna}
Eric Nguyen, Michael Poli, Marjan Faizi, Armin Thomas, Michael Wornow, Callum
  Birch-Sykes, Stefano Massaroli, Aman Patel, Clayton Rabideau, Yoshua Bengio,
  et~al.
\newblock Hyenadna: Long-range genomic sequence modeling at single nucleotide
  resolution.
\newblock \emph{Advances in neural information processing systems},
  36:\penalty0 43177--43201, 2023.

\bibitem[Nordheim \& Rich(1983)Nordheim and Rich]{nordheim1983negatively}
Alfred Nordheim and Alexander Rich.
\newblock Negatively supercoiled simian virus 40 dna contains z-dna segments
  within transcriptional enhancer sequences.
\newblock \emph{Nature}, 303\penalty0 (5919):\penalty0 674--679, 1983.

\bibitem[Panayotatos \& Wells(1981)Panayotatos and
  Wells]{panayotatos1981cruciform}
Nikos Panayotatos and Robert~D Wells.
\newblock Cruciform structures in supercoiled dna.
\newblock \emph{Nature}, 289\penalty0 (5797), 1981.

\bibitem[Papineni et~al.(2025)Papineni, Dutta, Chao, Acanto, Sathian, Surana,
  and Davuluri]{papineni2025augmenting}
Nimisha Papineni, Pratik Dutta, Max Chao, Orbin Acanto, Rekha Sathian, Pallavi
  Surana, and Ramana~V Davuluri.
\newblock Augmenting dnabert embeddings with multimodal dna features for
  improved regulatory sequence interpretation.
\newblock In \emph{MLCB}, pp.\  294--303, 2025.

\bibitem[SantaLucia~Jr(1998)]{santalucia1998unified}
John SantaLucia~Jr.
\newblock A unified view of polymer, dumbbell, and oligonucleotide dna
  nearest-neighbor thermodynamics.
\newblock \emph{Proceedings of the National Academy of Sciences}, 95\penalty0
  (4):\penalty0 1460--1465, 1998.

\bibitem[Schiff et~al.(2024)Schiff, Kao, Gokaslan, Dao, Gu, and
  Kuleshov]{schiff2024caduceus}
Yair Schiff, Chia-Hsiang Kao, Aaron Gokaslan, Tri Dao, Albert Gu, and Volodymyr
  Kuleshov.
\newblock Caduceus: Bi-directional equivariant long-range dna sequence
  modeling.
\newblock \emph{Proceedings of machine learning research}, 235:\penalty0 43632,
  2024.

\bibitem[Trop et~al.(2024)Trop, Schiff, Marroquin, Kao, Gokaslan, Polen, Shao,
  de~Almeida, Pierrot, Li, et~al.]{trop2024genomics}
Evan Trop, Yair Schiff, Edgar~Mariano Marroquin, Chia~Hsiang Kao, Aaron
  Gokaslan, McKinley Polen, Mingyi Shao, Bernardo~P de~Almeida, Thomas Pierrot,
  Yang~I Li, et~al.
\newblock The genomics long-range benchmark: advancing dna language models,
  2024.

\bibitem[Wang et~al.(1979)Wang, Quigley, Kolpak, Crawford, Van~Boom, Van
  Der~Marel, and Rich]{wang1979molecular}
Andrew H-J Wang, Gary~J Quigley, Francis~J Kolpak, James~L Crawford, Jacques~H
  Van~Boom, Gijs Van Der~Marel, and Alexander Rich.
\newblock Molecular structure of a left-handed double helical dna fragment at
  atomic resolution.
\newblock \emph{Nature}, 282\penalty0 (5740):\penalty0 680--686, 1979.

\bibitem[Yang et~al.(2024)Yang, Zhang, Zheng, Zhang, Liang, and
  Tang]{yang2024employing}
Minghao Yang, Shichen Zhang, Zhihang Zheng, Pengfei Zhang, Yan Liang, and
  Shaojun Tang.
\newblock Employing bimodal representations to predict dna bendability within a
  self-supervised pre-trained framework.
\newblock \emph{Nucleic Acids Research}, 52\penalty0 (6):\penalty0 e33--e33,
  2024.

\bibitem[Yang et~al.(2026)Yang, Liu, Cao, and Su]{yang2026d3lm}
Zhao Yang, Hengchang Liu, Chuan Cao, and Bing Su.
\newblock D3lm: A discrete dna diffusion language model for bidirectional dna
  understanding and generation.
\newblock \emph{arXiv preprint arXiv:2603.01780}, 2026.

\bibitem[Zhang et~al.(2023)Zhang, Zhang, He, Zhang, Qin, and
  Yao]{zhang2023dnagpt}
Daoan Zhang, Weitong Zhang, Bing He, Jianguo Zhang, Chenchen Qin, and Jianhua
  Yao.
\newblock Dnagpt: a generalized pretrained tool for multiple dna sequence
  analysis tasks.
\newblock \emph{BioRxiv}, pp.\  2023--07, 2023.

\bibitem[Zhou \& Troyanskaya(2015)Zhou and Troyanskaya]{zhou2015predicting}
Jian Zhou and Olga~G Troyanskaya.
\newblock Predicting effects of noncoding variants with deep learning--based
  sequence model.
\newblock \emph{Nature methods}, 12\penalty0 (10):\penalty0 931--934, 2015.

\bibitem[Zhou et~al.(2013)Zhou, Yang, Lu, Dror, Dantas~Machado, Ghane,
  Di~Felice, and Rohs]{zhou2013dnashape}
Tianyin Zhou, Lin Yang, Yan Lu, Iris Dror, Ana~Carolina Dantas~Machado, Tahereh
  Ghane, Rosa Di~Felice, and Remo Rohs.
\newblock Dnashape: a method for the high-throughput prediction of dna
  structural features on a genomic scale.
\newblock \emph{Nucleic acids research}, 41\penalty0 (W1):\penalty0 W56--W62,
  2013.

\bibitem[Zhou et~al.(2015)Zhou, Shen, Yang, Abe, Horton, Mann, Bussemaker,
  Gord{\^a}n, and Rohs]{zhou2015quantitative}
Tianyin Zhou, Ning Shen, Lin Yang, Namiko Abe, John Horton, Richard~S Mann,
  Harmen~J Bussemaker, Raluca Gord{\^a}n, and Remo Rohs.
\newblock Quantitative modeling of transcription factor binding specificities
  using dna shape.
\newblock \emph{Proceedings of the National Academy of Sciences}, 112\penalty0
  (15):\penalty0 4654--4659, 2015.

\bibitem[Zhou et~al.(2024)Zhou, Ji, Li, Dutta, Davuluri, and
  Liu]{zhou2024dnabert}
Zhihan Zhou, Yanrong Ji, Weijian Li, Pratik Dutta, Ramana Davuluri, and Han
  Liu.
\newblock Dnabert-2: Efficient foundation model and benchmark for multi-species
  genomes.
\newblock In \emph{International Conference on Learning Representations},
  volume 2024, pp.\  41642--41665, 2024.

\end{thebibliography}

%%%%%%%%%%%%%%%%%%%%%%%%%%%%%%%%%%%%%%%%%%%%%%%%%%%%%%%%%%%%
% Appendix — must appear AFTER the bibliography (ICLR requirement).
% Appendix pages do not count toward the main-text page limit.
%%%%%%%%%%%%%%%%%%%%%%%%%%%%%%%%%%%%%%%%%%%%%%%%%%%%%%%%%%%%

\appendix

\section{Local Molecular Descriptor Definitions}
\label{app:local-features}

This section defines the sequence-derived molecular descriptors used as local features in VANDAM. 
We specify the 14 DNAshapeR channels used by the local-feature-injection pathway.
The mechanism used to incorporate these descriptors into the backbone is defined separately in \S\ref{sec:lfi}. Table~\ref{tab:dnashape-catalog} summarizes the 14 DNAshapeR local descriptors.

\begin{table}[H]
\centering
\caption{The 14 DNAshapeR channels used by VANDAM. Rotational quantities are measured in degrees and translational quantities in \AA.}
\label{tab:dnashape-catalog}
\scriptsize
\setlength{\tabcolsep}{3pt}
\begin{tabular}{p{0.08\textwidth}p{0.15\textwidth}cp{0.20\textwidth}p{0.40\textwidth}}
\toprule
\textbf{Feature} & \textbf{Full name} & \textbf{Units} & \textbf{Structural category} & \textbf{Biophysical interpretation} \\
\midrule
MGW & Minor Groove Width & \AA & Groove geometry & Width of the minor groove at the center base; narrow grooves deepen the local negative electrostatic potential. \\
HelT & Helical Twist & degrees & Inter-bp rotation & Rotation about the helix axis between successive base-pair steps. \\
ProT & Propeller Twist & degrees & Intra-bp rotation & Out-of-plane twist of the two bases within a pair. \\
Roll & Roll & degrees & Inter-bp rotation & Rotation about the long axis of a base-pair step; a principal component of intrinsic bending. \\
EP & Electrostatic Potential & $kT/e$ & Groove/electrostatic & Electrostatic potential in the minor groove at the center base. \\
Stretch & Stretch & \AA & Intra-bp translation & Separation of the two bases along the base-pair long axis. \\
Tilt & Tilt & degrees & Inter-bp rotation & Rotation about the short axis of a base-pair step; the second bending component used with Roll. \\
Buckle & Buckle & degrees & Intra-bp rotation & Out-of-plane bending of the two bases toward the same groove. \\
Shear & Shear & \AA & Intra-bp translation & Lateral displacement of the two bases along the short axis. \\
Opening & Opening & degrees & Intra-bp rotation & Opening of the base pair toward the major groove. \\
Rise & Rise & \AA & Inter-bp translation & Distance along the helix axis between successive base pairs. \\
Shift & Shift & \AA & Inter-bp translation & Displacement of the next base pair along the short axis. \\
Stagger & Stagger & \AA & Intra-bp translation & Displacement of the two bases along the long axis relative to each other. \\
Slide & Slide & \AA & Inter-bp translation & Displacement of the next base pair along the long axis. \\
\bottomrule
\end{tabular}
\end{table}

\section{Local-Feature-Injection Alignment and Backbone-Specific Handling}
\label{app:local-feature-injection}

This section specifies the visibility rules and backbone adaptations for the input-concatenation injector of \S\ref{sec:lfi}.

\paragraph{Visibility and warm-up.}
Features are computed from the corrupted sequence available to the backbone, never from the uncorrupted target sequence. 
For MLM, a masked nucleotide is excluded both at its own position and from the feature windows of neighbouring positions. 
For NTP, feature windows exclude the target and all future nucleotides. 
Special tokens receive zero features, and no feature vectors are learned for them. 
The backbone is frozen while the feature projection $W_f$ and local-feature-injection projection $W_m$ are initialized against its fixed representation; all parameters are then trained jointly.

\paragraph{BPE token alignment.}
The features operate at nucleotide resolution, whereas a DNABERT-2 BPE token can span multiple bases. We compute features at base resolution and mean-pool them over the bases spanned by each token. Computing features before pooling preserves context that crosses BPE-token boundaries. Special tokens receive zero features.

\paragraph{Causal alignment.}
For causal HyenaDNA, the centered pentamer $(x_{i-2},\ldots,x_{i+2})$ cannot be used at position $i$, because it includes future bases. We instead use the right-aligned window $(x_{i-4},\ldots,x_i)$, which contains only bases $x_j$ with $j \leq i$. Thus, the adaptation changes the feature's determining window, not the position at which it is injected.

\paragraph{Backbone-specific handling.}
Caduceus-PS represents forward and reverse-complement strands jointly, so VANDAM applies the same local-feature-injection projection $W_m$ independently to each strand in its natural orientation before restoring the paired representation. This preserves reverse-complement equivariance.

\section{Regional Biological Feature Definitions}
\label{app:regional-features}

We define the six regional properties of \S\ref{sec:rpp} for a region $\mathcal{T}=(x_s,\ldots,x_{s+K-1})$ of length $K$. These properties are computed directly from the nucleotide sequence and summarize regional geometry, composition, duplex stability, and non-B-DNA structural propensity.

Two of the regional properties use additional sequence-derived quantities that are not included among the local molecular features supplied to LFI: the SantaLucia nearest-neighbor duplex free energy and the purine--pyrimidine classification of each nucleotide. These quantities are introduced below only as intermediate inputs to the melting-propensity and Z-DNA-propensity definitions; they are not separate RPP targets.

\paragraph{Nearest-neighbor duplex free energy.}
For a valid dinucleotide $XY$, let $\Delta G^{\circ}_{37}(X,Y)$ denote its sequence-dependent nearest-neighbor free-energy contribution at $37\,^{\circ}\mathrm{C}$. We use the SantaLucia values listed in Table~\ref{tab:dinucleotide-dg} \citep{santalucia1998unified}. Steps related by Watson--Crick reverse-complement symmetry share the same value.

\begin{table}[H]
\centering
\caption{SantaLucia nearest-neighbor duplex free energies at $310.15\,\mathrm{K}$ ($37\,^{\circ}\mathrm{C}$) \citep{santalucia1998unified}.}
\label{tab:dinucleotide-dg}
\small
\begin{tabular}{lc@{\hspace{28pt}}lc}
\toprule
\textbf{Step (5'--3')} & \textbf{Free energy (kcal/mol)} & \textbf{Step (5'--3')} & \textbf{Free energy (kcal/mol)} \\
\midrule
AA / TT & $-1.00$ & CT / AG & $-1.28$ \\
AT      & $-0.88$ & GA / TC & $-1.30$ \\
TA      & $-0.58$ & CG      & $-2.17$ \\
CA / TG & $-1.45$ & GC      & $-2.24$ \\
GT / AC & $-1.44$ & GG / CC & $-1.84$ \\
\bottomrule
\end{tabular}
\end{table}

\paragraph{Purine--pyrimidine classification.}
For each valid nucleotide $b\in\{A,C,G,T\}$, define
\begin{equation}
\operatorname{class}(b)=
\begin{cases}
\mathrm{purine}, & b\in\{A,G\},\\
\mathrm{pyrimidine}, & b\in\{C,T\}.
\end{cases}
\end{equation}
This classification is used below to identify alternating purine--pyrimidine tracts associated with Z-DNA formation.

\paragraph{Intrinsic curvature (dimensionless).}
Let $\rho_k$, $\tau_k$, and $\theta_k$ be the raw DNAshapeR Roll, Tilt, and HelT values at position $k$, converted from degrees to radians. Starting from the identity frame $R_{-1}=I_3$, we propagate
\begin{equation}
Q_k=R_z(\theta_k)R_y(\rho_k)R_x(\tau_k),\qquad
R_k=R_{k-1}Q_k,\qquad
t_k=R_ke_3,
\end{equation}
for every $k\in\{0,\ldots,K-1\}$, including positions whose boundary DNAshape values are zero. The curvature score is
\begin{equation}
c(\mathcal{T})=\frac{1}{K}\left\lVert\sum_{k=0}^{K-1}t_k\right\rVert_2.
\end{equation}
It is the end-to-end/contour-length ratio: values near one indicate aligned tangents, whereas lower values indicate stronger intrinsic curvature. The score captures whether sequence-dependent local bends reinforce coherently across the region \citep{zhou2013dnashape,chiu2016dnashaper}.

\paragraph{AT content (dimensionless fraction).}
Let $N_b$ denote the number of occurrences of base $b$, and let $N_{\mathrm{valid}}=N_A+N_C+N_G+N_T$. The implemented score is
\begin{equation}
a(\mathcal{T})=1-\frac{N_C+N_G}{\max(N_{\mathrm{valid}},1)}.
\end{equation}
Non-ACGT positions contribute to neither the numerator nor the valid-base count. This property captures AT-rich regional composition, a component of bendable and low-stability origin-like sequence contexts.

\paragraph{Melting propensity (kcal/mol).}
Let $d_0=0$ and, for $i\geq1$, define
\begin{equation}
d_i=
\begin{cases}
\Delta G^{\circ}_{37}(x_{s+i-1},x_{s+i}), & x_{s+i-1},x_{s+i}\in\{A,C,G,T\},\\
0, & \text{otherwise}.
\end{cases}
\end{equation}
Thus, $(d_0,\ldots,d_{K-1})$ is a position-aligned track of nearest-neighbor free-energy contributions, with a leading zero because the first nucleotide has no preceding nucleotide within the region. Using the SantaLucia values in Table~\ref{tab:dinucleotide-dg}, the regional score is
\begin{equation}
m(\mathcal{T})=\frac{1}{K-1}\sum_{i=0}^{K-1}d_i
=\frac{1}{K-1}\sum_{i=1}^{K-1}d_i.
\end{equation}
More negative values indicate a more stable duplex, whereas less negative values indicate easier strand separation \citep{santalucia1998unified}.

\paragraph{G-quadruplex propensity (dimensionless).}
We follow the standard G4Hunter definition, which caps run scores at 4 and uses a default 25-nt window \citep{bedrat2016re}. For each position $i$, let $\ell_i$ be the length of the maximal contiguous run of the same base containing $i$. Define the per-base G4Hunter track
\begin{equation}
g_i=
\begin{cases}
\min(\ell_i,4), & x_{s+i}=G,\\
-\min(\ell_i,4), & x_{s+i}=C,\\
0, & \text{otherwise},
\end{cases}
\end{equation}
where A, T, and non-ACGT positions receive zero and terminate G/C runs. For $K>0$, set $w=\min(25,K)$ and compute
\begin{equation}
q(\mathcal{T})=\max_{0\leq j\leq K-w}
\left|\frac{1}{w}\sum_{i=j}^{j+w-1}g_i\right|;
\qquad q(\emptyset)=0.
\end{equation}
Thus, regions shorter than 25 nucleotides are evaluated using one window spanning the complete region. High values identify clustered G- or C-runs with G-quadruplex potential on either strand \citep{bedrat2016re}.

\paragraph{Cruciform propensity (bp).}
Let $\overline{x}$ denote the Watson--Crick complement of base $x$. We define the one-arm stem length
\begin{equation}
\ell_{\mathrm{stem}}(\mathcal{T})
=
\max\left\{\ell:
\begin{array}{l}
\exists\,i<j,\quad 3\leq j-i-1\leq12,\\
x_{s+i-r}=\overline{x_{s+j+r}}\quad\text{for all }r=0,\ldots,\ell-1
\end{array}
\right\}.
\end{equation}
The indices are restricted to $\mathcal{T}$, and the score has a default value of zero when no admissible stem exists. Both bases in every pair must be valid ACGT; an invalid base fails the complement match and terminates the stem. The score measures one arm rather than the total $2\ell$ paired bases. Long inverted-repeat stems can extrude as cruciform or hairpin structures under supercoiling and are associated with recombination and genomic instability \citep{panayotatos1981cruciform,brazda2011cruciform}.

\paragraph{Z-DNA propensity (bp).}
For $i=0,\ldots,K-2$, define
\begin{equation}
u_i=
\begin{cases}
1, & x_{s+i},x_{s+i+1}\in\{A,C,G,T\}\ \text{and}\
\operatorname{class}(x_{s+i})\neq\operatorname{class}(x_{s+i+1}),\\
0, & \text{otherwise}.
\end{cases}
\end{equation}
If $r(u)$ is the length of the longest consecutive run of ones in $(u_0,\ldots,u_{K-2})$, then
\begin{equation}
z_{\mathrm{DNA}}(\mathcal{T})=r(u)+1.
\end{equation}
An invalid base makes both adjacent alternation indicators zero and therefore terminates a tract; a valid homopolymer has score one. The score therefore measures the length of an alternating purine--pyrimidine tract, a sequence class associated with left-handed Z-DNA formation, particularly under negative supercoiling \citep{wang1979molecular,nordheim1983negatively}. It is a coarse sequence proxy and does not model sequence-dependent Z-DNA stability.

\section{RPP Implementation and Backbone Alignment}
\label{app:rpp-implementation}

The main text defines the shared prediction head, RPP loss, and loss schedule. Here we give the remaining implementation details. For each base-defined tile, the regional-property target vector is formed using target-specific transformations and calibration statistics estimated offline from training tiles.

\paragraph{Tiling and pooling.}
Targets are computed only from bases within the crop; no flanking sequence outside the crop is consulted. Tiles with insufficient valid A/C/G/T content are excluded from the RPP loss. For a tile $\mathcal{T}_j$ of length $K$, let
$N_{\mathrm{valid}}(\mathcal{T}_j)$ be its number of A/C/G/T bases. We define
the set of valid tiles as
\begin{equation}
\mathcal{V}
=
\left\{
j :
\frac{N_{\mathrm{valid}}(\mathcal{T}_j)}{K}
\geq 0.5
\right\}.
\end{equation}
Tiles with fewer than 50\% valid A/C/G/T bases are excluded from the RPP loss.

For a valid tile, its hidden states are aggregated by scalar-attention pooling,
\begin{equation}
    \alpha_i =
    \frac{\exp(w^\top H_i)}
         {\sum_{k\in\mathcal{T}_t}\exp(w^\top H_k)},
    \qquad
    z_t = \sum_{i\in\mathcal{T}_t}\alpha_i H_i,
\end{equation}
where the attention weights are normalized independently within each tile. The pooling scorer is shared across target dimensions.

\paragraph{BPE alignment.}
For DNABERT-2, tile boundaries and targets remain defined in nucleotide coordinates rather than BPE-token counts. We reconstruct the retained nucleotide sequence by decoding BPE tokens and assign each nonempty token to one tile according to the midpoint of its decoded nucleotide span. A token crossing a tile boundary is assigned wholly to the tile containing its midpoint; it is neither split nor duplicated. Special, padding, and non-nucleotide tokens are excluded from pooling. Consequently, all backbones are supervised against the same type of base-defined regional targets while their native token states are pooled according to their respective tokenizations.

\paragraph{Backbone-specific hyperparameters.}
Table~\ref{tab:rpp-hyperparams} reports the tile length and $\lambda_{\max}$ used for the reported VANDAM-PT configuration of each backbone. DNABERT-2 uses a 150\,bp tile rather than the 300\,bp used elsewhere, reflecting its shorter native context length (Table~\ref{tab:models}). $\lambda_{\max}$ is set per backbone to hold the ratio between the native token loss and $\mathcal{L}_{\mathrm{RPP}}$ approximately constant, rather than fixed to a single value across backbones.

\begin{table}[h]
\centering
\caption{RPP tile length and loss-weighting coefficient used for the reported VANDAM-PT configuration of each backbone.}
\label{tab:rpp-hyperparams}
\small
\begin{tabular}{lcc}
\toprule
\textbf{Backbone} & \textbf{Tile length (bp)} & $\boldsymbol{\lambda_{\max}}$ \\
\midrule
NTv3-8M     & 300 & 0.5 \\
Caduceus    & 300 & 0.5 \\
DNABERT-2   & 150 & 1.0 \\
HyenaDNA    & 300 & 0.1 \\
\bottomrule
\end{tabular}
\end{table}

\section{Training Configuration}
\label{app:training-details}

This section reports the training setup used across backbones.
Both regimes continue training from each backbone's released checkpoint.
Within each regime, the VANDAM arms use the settings below unchanged and differ only in the biological components under study, so the comparisons in Section~\ref{sec:setup} hold data, optimizer, schedule and number of updates fixed.

\paragraph{Checkpoint sources.}
We initialize from \emph{InstaDeepAI/NTv3\_8M\_pre} (NTv3-8M), \emph{kuleshov-group/caduceus-ps\_seqlen-131k\_d\_model-256\_n\_layer-16} (Caduceus-PS), \emph{zhihan1996/DNABERT-2-117M} (DNABERT-2-117M), and \emph{LongSafari/hyenadna-small-32k-seqlen-hf} (HyenaDNA-small-32k). We use the shorter family names Caduceus and HyenaDNA elsewhere in the paper.

\begin{table}[h]
\centering
\caption{Training configuration used across backbones.}
\label{tab:training-config}
\small
\begin{tabular}{ll}
\toprule
\textbf{Parameter} & \textbf{Value} \\
\midrule
Initialization        & Corresponding released backbone checkpoint (all parameters trainable) \\
Optimizer steps       & 54{,}000 \\
Global batch size     & 128 sequences \\
Peak learning rate    & $6 \times 10^{-4}$ \\
Schedule              & cosine, 600 warmup steps, decayed to zero at the final step \\
Optimizer             & AdamW, $\beta = (0.9, 0.999)$, weight decay $0.01$, gradient clipping $1.0$ \\
Precision             & bf16 \\
\bottomrule
\end{tabular}
\end{table}

\begin{table}[h]
\centering
\caption{Per-backbone compute, used both in Baseline and VANDAM.}
\label{tab:per-backbone-compute}
\small
\begin{tabular}{lcc}
\toprule
\textbf{Backbone} & \textbf{Nucleotides / tokens seen} & \textbf{Total compute} \\
\midrule
NTv3-8M        & 38.221B nucleotides & 1.764 EFLOP \\
Caduceus-PS       & 56.623B nucleotides & 2.623 EFLOP \\
DNABERT-2-117M    & 21.234B source nucleotides; 3.539B BPE tokens & 2.419 EFLOP \\
HyenaDNA-small-32k & 56.623B nucleotides & 1.114 EFLOP \\
\bottomrule
\end{tabular}
\end{table}

\paragraph{Data and augmentation.}
Training uses OpenGenome2. NTv3 uses a mixed-length curriculum: sequence lengths of 1{,}024, 2{,}048, 4{,}096 and 8{,}192\,bp are sampled with probabilities $0.1$, $0.15$, $0.25$ and $0.5$, giving a weighted mean of 5{,}530\,bp per example. Caduceus and HyenaDNA instead use a fixed 8{,}192\,bp context. DNABERT-2 uses a fixed 3{,}072-bp crop, tokenized to a fixed 512-BPE-token input including [CLS] and [SEP]; longer tokenizations are truncated and shorter ones padded.
Evaluation uses the full 8{,}192\,bp context for NTv3, Caduceus, and HyenaDNA. DNABERT-2 is evaluated on 2{,}500-bp crops.
Two augmentations are applied during training only: a random crop of the target length from each record, and reverse-complementation with probability $0.5$.
Evaluation is forward-strand.

\paragraph{Native token objectives.}
NTv3, Caduceus, and DNABERT-2 use standard random per-token masking at a rate of 15\%, following the usual 80/10/10 split between mask, random, and unchanged tokens. HyenaDNA retains its native autoregressive next-token prediction objective.

\paragraph{Multi-task schedule.}
Every tenth optimizer step is a task-only update. We select one task from the pool in Appendix~\ref{app:pool} in deterministic round-robin order and optimize only its supervised task loss; the native sequence objective and, for VANDAM-MT, the RPP losses are not evaluated on these steps. All remaining updates are sequence-only: Baseline-MT optimizes the native sequence objective, whereas VANDAM-MT optimizes the native sequence objective together with the shared RPP losses, with unit weight for each active loss term. Both Baseline-MT and VANDAM-MT include a task-specific supervised head for each supervised pool task; VANDAM-MT additionally includes the shared RPP head, which is updated only on sequence-only batches.

For the first 1{,}000 steps, the backbone is frozen during task-only updates so that the task heads calibrate against a stable representation; it remains trainable during sequence-only updates. Task batches use the full 8{,}192\,bp context, except for DNABERT-2, which uses its fixed 3{,}072-bp, 512-BPE-token training configuration. For the two zero-shot variant tasks, we do not use a supervised task head; instead, we score variants using the sequence log-likelihood ratio between the reference and alternate alleles.

\section{Multi-Task Training Pool and Overlap Control}
\label{app:pool}

The multi-task pool contains eight supervised tasks in three families:
single-sequence classification, paired variant-effect classification, and
zero-shot SNV scoring. 

\begin{table}[h]
\centering
\caption{The eight-task multi-task training pool. The training tasks are
task-disjoint from the evaluation suite; record-level filtering further removes
training examples that overlap an evaluation example under the policy below.}
\label{tab:mt-pool}
\small
\begin{tabular}{ll}
\toprule
\textbf{Family} & \textbf{Tasks} \\
\midrule
Single-sequence classification & promoter detection, splice-site detection, enhancer annotation \\
Variant-effect classification (paired) & common-vs-rare, meQTL, sQTL \\
Zero-shot SNV scoring & BEND VEP disease, BEND VEP expression \\
\bottomrule
\end{tabular}
\end{table}

All reported multi-task results use a single strict retained training pool. We compared every candidate training example with every available split of each of the nine evaluation tasks. The nine tasks comprise 16 evaluation task/split partitions, yielding 128 training--evaluation-split comparisons. For coordinate-resolvable records, coordinates were harmonized to GRCh38, and GRCh37/hg19 records were lifted to GRCh38 before comparison. For sequence-only tasks, we assigned a genomic locus only when an exact, full-length alignment was unique. Training examples with ambiguous, unmapped, or unliftable loci were excluded rather than assigned arbitrary coordinates.

The strict filter excluded a training example if it shared with any evaluation example: (i) a normalized allele, (ii) a genomic position irrespective of alternate allele, (iii) an overlapping labelled anchor interval, (iv) an identical model-input sequence, treating reverse complements as equal, (v) complete containment of the shorter labelled or model-input sequence, (vi) overlap between the two loci after expanding each locus by 4{,}096 bp on each flank, or (vii) overlap between the corresponding model-input windows. Accordingly, for coordinate-resolvable records, retained multi-task training examples are locus/window-disjoint from evaluation examples under these declared rules. Table~\ref{tab:mt-filter} reports the resulting training pool. 

\begin{table}[t]
\centering
\caption{Record-level filtering of the eight-task multi-task training pool.
Counts are examples before filtering; the retained pool was
used for all reported multi-task results.}
\label{tab:mt-filter}
\small
\begin{tabular}{lrrr}
\toprule
\textbf{Training task} & \textbf{Visible} & \textbf{Excluded} & \textbf{Retained} \\
\midrule
GUE promoter                 & 47{,}356    & 40{,}183 (84.85\%) & 7{,}173 \\
GUE splice site              & 36{,}496    & 18{,}006 (49.34\%) & 18{,}490 \\
LRB enhancer                 & 99{,}995    & 28{,}987 (28.99\%) & 71{,}008 \\
Common-vs-rare               & 193{,}908   & 50{,}620 (26.11\%) & 143{,}288 \\
meQTL                        & 73{,}601    & 25{,}108 (34.11\%) & 48{,}493 \\
sQTL                         & 1{,}000{,}741 & 385{,}758 (38.55\%) & 614{,}983 \\
BEND VEP disease            & 295{,}495   & 154{,}346 (52.23\%) & 141{,}149 \\
BEND VEP expression         & 105{,}263   & 38{,}825 (36.88\%) & 66{,}438 \\
\midrule
\textbf{Total}              & \textbf{1{,}852{,}855} & \textbf{741{,}833 (40.04\%)} & \textbf{1{,}111{,}022} \\
\bottomrule
\end{tabular}
\end{table}

\section{Evaluation Suite}
\label{app:eval-tasks}

\subsection{Tasks}

Table~\ref{tab:eval-suite} summarizes the nine evaluation tasks.

\begin{table}[h]
\centering
\caption{The nine-task evaluation suite. Len denotes the task-level sequence window. NTv3, Caduceus, and HyenaDNA use the listed context; DNABERT-2 uses 2{,}500-bp crops.}
\label{tab:eval-suite}
\scriptsize
\begin{tabular}{llllrr}
\toprule
\textbf{Task} & \textbf{Access} & \textbf{Prediction} & \textbf{Metric} & \textbf{Test $n$} & \textbf{Len (bp)} \\
\midrule
BRCA1 variant effect             & zero-shot    & cosine dissimilarity & AUROC              & 3{,}893   & 8{,}192 \\
Clinical pathogenicity (ClinVar) & zero-shot    & cosine dissimilarity & AUROC              & 258{,}944 & 8{,}192 \\
Disease variants (OMIM)          & zero-shot    & cosine dissimilarity & AUROC              & 200{,}406 & 8{,}192 \\
Mendelian variants (TraitGym)    & zero-shot    & cosine dissimilarity & AUROC              & 3{,}380   & 8{,}192 \\
\midrule
Transcription-factor binding     & linear probe & probability          & AUROC              & 5{,}000   & 100 \\
Regulatory variants (eQTL)       & linear probe & probability          & AUROC              & 8{,}862   & 8{,}192 \\
Gene-expression variants         & linear probe & probability          & AUROC              & 9{,}147   & 1{,}023 \\
Chromatin state (histone)        & linear probe & 20 outputs           & macro AUROC        & 227{,}456 & 8{,}192 \\
CAGE expression                  & linear probe & regression           & macro Pearson $r$  & 1{,}922   & 8{,}192 \\
\bottomrule
\end{tabular}
\end{table}

\paragraph{Zero-shot tasks.}
BRCA1 uses saturation genome-editing data, distinguishing functionally impaired from functional single-nucleotide variants \citep{findlay2018accurate}.
ClinVar uses the VEP-eval benchmark and asks for pathogenic versus benign classification \citep{lu2025genomic}.
The OMIM task originates from LRB and asks for pathogenic non-coding regulatory variants against common background variation. Our GFMBench evaluation retains all 406 positives and uses 200{,}000 background variants, giving 200{,}406 examples \citep{trop2024genomics,larey2026gfmbench}.
TraitGym asks for causal non-coding variants for Mendelian traits against matched controls \citep{benegas2025benchmarking}.

\paragraph{Supervised tasks.}
Transcription-factor binding is binary classification on short sequences, the only task in the suite running at 100\,bp \citep{zhou2024dnabert}.
The eQTL task predicts whether a variant is causal for nearby gene expression, conditioned on tissue \citep{trop2024genomics}.
Gene-expression variants uses the expression-effect subset of the BioFM Variant Benchmark, derived from DeepSEA regulatory annotations \citep{medvedev2025biotoken,zhou2015predicting}.
Histone marks follows LRB's 20-track multi-label task over annotated 200\,bp bins, derived from the DeepSEA chromatin-feature dataset \citep{trop2024genomics,zhou2015predicting}.
CAGE follows LRB's binned-expression task across 50 FANTOM5 tracks in log space, constructed from Basenji data \citep{trop2024genomics,kelley2018sequential}.

\subsection{Prediction Protocols}

Each task is evaluated by first producing a score or prediction and then applying the appropriate evaluation metric. We describe these two steps separately below.

\paragraph{Zero-shot variant scoring.}
Reference and alternate sequences are encoded independently using each backbone's native tokenizer.
Their final contextualized token states are pooled by an attention-masked mean to obtain sequence representations $h_{\mathrm{ref}}$ and $h_{\mathrm{alt}}$.
We score each variant by cosine dissimilarity, $s_i = 1 - \cos(h_{\mathrm{alt}}^{(i)}, h_{\mathrm{ref}}^{(i)})$, so that a larger representational perturbation ranks as more likely to be pathogenic.
Nothing is fitted on the evaluation task.
Although each sequence pair differs at only one nucleotide, the resulting representation change is not necessarily confined to one hidden state.
In single-nucleotide models, the perturbation can propagate across contextualized positions through convolution, attention, or state-space mixing; in DNABERT-2, it may additionally alter the local BPE segmentation.
The number and magnitude of affected hidden states are therefore backbone-dependent.
Because the reference and alternate sequences share nearly all of their genomic context and are summarized by global mean pooling, however, their sequence-level representations are generally similar.

\paragraph{Linear probing.}
For supervised tasks, the backbone is frozen and a single linear head is trained for three epochs with AdamW at learning rate $3 \times 10^{-5}$, weight decay $0.01$, and batch size 16. For binary single-sequence tasks, the head maps a pooled representation to a positive-class probability. For the two paired variant-effect tasks, it consumes $\left[
\,h_{\mathrm{alt}}\mathbin{\Vert}h_{\mathrm{ref}}\,
\right]$, allowing an asymmetric comparison. The histone-marks head emits twenty independent sigmoid outputs trained with binary cross-entropy, and the CAGE head maps bin-pooled states to predicted $\log 1p$ counts using MSE.

\subsection{Evaluation Metrics}

\paragraph{AUROC.}
AUROC is used for the four zero-shot variant tasks and the binary linear-probe tasks. For the former, it evaluates whether cosine dissimilarity ranks pathogenic above benign variants within each backbone; absolute cosine distances are not compared across backbones. For the latter, it is computed from the predicted positive-class probability.

\paragraph{Macro AUROC.}
For histone marks, AUROC is computed independently for each track and then averaged, excluding tracks with only one class present in the test set. Thus, each track contributes equally to the reported score.

\paragraph{Macro Pearson correlation.}
For CAGE, all (sequence, bin) pairs are flattened separately for each track, Pearson $r$ is computed per track, and the mean over the fifty tracks is reported.
Chance is zero rather than $0.5$, which is why this task is never folded into the AUROC mean.

\section{Paired Statistical Inference for Main Results}
\label{app:paired-inference}

This appendix gives the inferential procedure used for Tables~\ref{tab:pretrain-pertask} and~\ref{tab:multitask}. For each task--backbone comparison, we estimate the paired score difference $\Delta=\mathrm{VANDAM}-\mathrm{Baseline}$, reported in the 0--100 units used by the main tables.

For the four zero-shot tasks and the two variant-effect probe tasks (eQTL and gene-expression variants), a test unit is one labeled reference--alternate variant pair, including the tissue covariate for eQTL. For TF binding, it is one labeled test sequence. For histone marks, it is one source sequence with all 20 track labels; for CAGE, it is one source region with all contiguous output bins across the 50 tracks. For all nine tasks, we construct 95\% confidence intervals using 10,000 paired bootstrap replicates that resample these test units with replacement, using the same selected units for VANDAM and Baseline. For the five frozen-probe tasks, each replicate additionally resamples the five matched probe seeds with replacement and recomputes the mean paired difference across the selected seeds. For the four zero-shot AUROC tasks, no probe is trained, so bootstrap replicates resample test units only. We use 10,000 paired randomization replicates for one-sided tests, swapping VANDAM and Baseline predictions within each test unit; for frozen-probe tasks, each swap is shared across seeds. Bootstrap resampling and paired randomization swaps preserve these test units intact. Benjamini--Hochberg adjustment is applied separately to the 36 comparisons in each regime and direction. A decision requires the corresponding adjusted one-sided $q<0.05$ and a 95\% paired interval wholly in that direction.

Tables~\ref{tab:pretrain-inference} and~\ref{tab:multitask-inference} report the complete results. The $q$ column is the adjusted value for the displayed supported direction; it is omitted when neither direction meets the decision rule.

\begin{table}[p]
\centering
\caption{Paired inference for pre-training comparisons. $\Delta$ is VANDAM minus Baseline in points; brackets give the paired 95\% confidence interval. The $q$ column reports the BH-adjusted one-sided $q$-value in the direction of the observed $\Delta$. ``Supported'' denotes the direction meeting the paired-interval and BH-adjusted $q<0.05$ criteria.}
\label{tab:pretrain-inference}
\scriptsize
\setlength{\tabcolsep}{3pt}
\begin{tabular}{llccc}
\toprule
\textbf{Backbone} & \textbf{Task} & \textbf{$\Delta$ [95\% CI]} & \textbf{$q$} & \textbf{Supported} \\
\midrule
NTv3-8M & ClinVar & $3.6$ [$3.3,4.0$] & $<0.001$ & VANDAM \\
NTv3-8M & OMIM & $9.3$ [$5.4,13.1$] & $<0.001$ & VANDAM \\
NTv3-8M & TraitGym & $6.1$ [$1.4,10.7$] & 0.008 & VANDAM \\
NTv3-8M & BRCA1 & $0.6$ [$-2.1,3.3$] & 0.425 & -- \\
NTv3-8M & TF binding & $1.1$ [$0.4,1.7$] & 0.001 & VANDAM \\
NTv3-8M & eQTL & $1.9$ [$1.4,2.3$] & $<0.001$ & VANDAM \\
NTv3-8M & Gene expression & $1.5$ [$0.9,2.1$] & $<0.001$ & VANDAM \\
NTv3-8M & Histone & $2.1$ [$1.9,2.3$] & $<0.001$ & VANDAM \\
NTv3-8M & CAGE & $10.7$ [$9.5,12.0$] & $<0.001$ & VANDAM \\
\midrule
Caduceus & ClinVar & $1.5$ [$1.2,1.8$] & $<0.001$ & VANDAM \\
Caduceus & OMIM & $6.3$ [$2.3,10.4$] & 0.002 & VANDAM \\
Caduceus & TraitGym & $8.8$ [$4.1,13.4$] & $<0.001$ & VANDAM \\
Caduceus & BRCA1 & $4.5$ [$1.5,7.4$] & 0.003 & VANDAM \\
Caduceus & TF binding & $6.3$ [$4.7,7.9$] & $<0.001$ & VANDAM \\
Caduceus & eQTL & $-0.5$ [$-0.7,-0.3$] & 0.653 & -- \\
Caduceus & Gene expression & $-0.6$ [$-1.1,-0.0$] & 0.282 & -- \\
Caduceus & Histone & $5.4$ [$5.3,5.6$] & $<0.001$ & VANDAM \\
Caduceus & CAGE & $-1.2$ [$-1.5,-0.9$] & 0.023 & Baseline \\
\midrule
DNABERT-2 & ClinVar & $6.6$ [$6.3,6.9$] & $<0.001$ & VANDAM \\
DNABERT-2 & OMIM & $-4.4$ [$-7.5,-1.4$] & 0.027 & Baseline \\
DNABERT-2 & TraitGym & $-0.5$ [$-4.0,3.1$] & 1.000 & -- \\
DNABERT-2 & BRCA1 & $0.7$ [$-1.7,3.2$] & 0.371 & -- \\
DNABERT-2 & TF binding & $1.3$ [$0.4,2.2$] & 0.003 & VANDAM \\
DNABERT-2 & eQTL & $2.8$ [$2.1,3.4$] & $<0.001$ & VANDAM \\
DNABERT-2 & Gene expression & $1.1$ [$0.2,2.0$] & 0.018 & VANDAM \\
DNABERT-2 & Histone & $3.8$ [$3.6,4.0$] & $<0.001$ & VANDAM \\
DNABERT-2 & CAGE & $0.3$ [$0.0,0.6$] & 0.044 & VANDAM \\
\midrule
HyenaDNA & ClinVar & $8.4$ [$8.2,8.6$] & $<0.001$ & VANDAM \\
HyenaDNA & OMIM & $-0.1$ [$-3.2,3.1$] & 1.000 & -- \\
HyenaDNA & TraitGym & $0.2$ [$-3.3,3.8$] & 0.557 & -- \\
HyenaDNA & BRCA1 & $3.4$ [$0.9,6.0$] & 0.007 & VANDAM \\
HyenaDNA & TF binding & $6.4$ [$5.3,7.4$] & $<0.001$ & VANDAM \\
HyenaDNA & eQTL & $1.1$ [$0.7,1.5$] & $<0.001$ & VANDAM \\
HyenaDNA & Gene expression & $0.4$ [$-0.2,1.0$] & 0.133 & -- \\
HyenaDNA & Histone & $-2.8$ [$-3.1,-2.5$] & 0.004 & Baseline \\
HyenaDNA & CAGE & $7.0$ [$6.2,7.7$] & $<0.001$ & VANDAM \\
\bottomrule
\end{tabular}
\end{table}

\begin{table}[p]
\centering
\caption{Paired inference for multi-task comparisons. $\Delta$ is VANDAM minus Baseline in points; brackets give the paired 95\% confidence interval. The $q$ column reports the BH-adjusted one-sided $q$-value in the direction of the observed $\Delta$. ``Supported'' denotes the direction meeting the paired-interval and BH-adjusted $q<0.05$ criteria.}
\label{tab:multitask-inference}
\scriptsize
\setlength{\tabcolsep}{3pt}
\begin{tabular}{llccc}
\toprule
\textbf{Backbone} & \textbf{Task} & \textbf{$\Delta$ [95\% CI]} & \textbf{$q$} & \textbf{Supported} \\
\midrule
NTv3-8M & ClinVar & $2.1$ [$1.9,2.4$] & $<0.001$ & VANDAM \\
NTv3-8M & OMIM & $2.5$ [$-1.0,6.0$] & 0.102 & -- \\
NTv3-8M & TraitGym & $5.3$ [$0.9,9.8$] & 0.015 & VANDAM \\
NTv3-8M & BRCA1 & $2.7$ [$-0.2,5.7$] & 0.051 & -- \\
NTv3-8M & TF binding & $1.1$ [$0.2,2.0$] & 0.010 & VANDAM \\
NTv3-8M & eQTL & $1.2$ [$0.7,1.6$] & $<0.001$ & VANDAM \\
NTv3-8M & Gene expression & $0.5$ [$-0.3,1.2$] & 0.124 & -- \\
NTv3-8M & Histone & $1.0$ [$0.9,1.1$] & $<0.001$ & VANDAM \\
NTv3-8M & CAGE & $4.2$ [$3.5,5.0$] & $<0.001$ & VANDAM \\
\midrule
Caduceus & ClinVar & $-3.0$ [$-3.3,-2.6$] & $<0.001$ & Baseline \\
Caduceus & OMIM & $5.7$ [$1.8,9.7$] & 0.004 & VANDAM \\
Caduceus & TraitGym & $7.9$ [$3.1,12.6$] & 0.001 & VANDAM \\
Caduceus & BRCA1 & $0.7$ [$-2.3,3.8$] & 0.343 & -- \\
Caduceus & TF binding & $12.7$ [$11.6,13.8$] & $<0.001$ & VANDAM \\
Caduceus & eQTL & $2.1$ [$1.4,2.9$] & $<0.001$ & VANDAM \\
Caduceus & Gene expression & $2.0$ [$0.8,3.3$] & 0.003 & VANDAM \\
Caduceus & Histone & $2.5$ [$2.2,2.9$] & $<0.001$ & VANDAM \\
Caduceus & CAGE & $0.9$ [$0.3,1.5$] & 0.163 & -- \\
\midrule
DNABERT-2 & ClinVar & $10.2$ [$9.9,10.4$] & $<0.001$ & VANDAM \\
DNABERT-2 & OMIM & $14.4$ [$11.2,17.7$] & $<0.001$ & VANDAM \\
DNABERT-2 & TraitGym & $2.5$ [$-1.3,6.3$] & 0.121 & -- \\
DNABERT-2 & BRCA1 & $3.8$ [$1.1,6.4$] & 0.005 & VANDAM \\
DNABERT-2 & TF binding & $4.5$ [$3.7,5.3$] & $<0.001$ & VANDAM \\
DNABERT-2 & eQTL & $3.0$ [$2.4,3.5$] & $<0.001$ & VANDAM \\
DNABERT-2 & Gene expression & $1.2$ [$0.3,2.1$] & 0.007 & VANDAM \\
DNABERT-2 & Histone & $5.4$ [$5.2,5.6$] & $<0.001$ & VANDAM \\
DNABERT-2 & CAGE & $-0.2$ [$-0.8,0.4$] & 1.000 & -- \\
\midrule
HyenaDNA & ClinVar & $15.5$ [$15.2,15.8$] & $<0.001$ & VANDAM \\
HyenaDNA & OMIM & $3.1$ [$-0.5,6.6$] & 0.064 & -- \\
HyenaDNA & TraitGym & $3.4$ [$-0.9,7.7$] & 0.084 & -- \\
HyenaDNA & BRCA1 & $1.9$ [$-1.1,4.9$] & 0.129 & -- \\
HyenaDNA & TF binding & $5.8$ [$4.8,6.8$] & $<0.001$ & VANDAM \\
HyenaDNA & eQTL & $0.3$ [$-0.1,0.6$] & 0.083 & -- \\
HyenaDNA & Gene expression & $-0.9$ [$-1.7,-0.1$] & 0.286 & -- \\
HyenaDNA & Histone & $1.0$ [$0.9,1.2$] & $<0.001$ & VANDAM \\
HyenaDNA & CAGE & $10.4$ [$9.2,11.5$] & $<0.001$ & VANDAM \\
\bottomrule
\end{tabular}
\end{table}

\clearpage
\section{Masked Feature Modeling and the Locality Control}
\label{app:mfm}

MFM predicts the local feature vector $\phi(x)_i$ from the backbone hidden state at each masked index $i \in \mathcal{M}$ using a thin head $g_{\mathrm{MFM}}$:
\begin{equation}
\mathcal{L}_{\mathrm{MFM}}
=
\frac{1}{|\mathcal{M}|}
\sum_{i \in \mathcal{M}}
\left\lVert g_{\mathrm{MFM}}(H_i)-\phi(x)_i \right\rVert_2^2 .
\end{equation}
For MFM (bio), $\phi(x)_i$ is the 15-dimensional local feature vector obtained by concatenating the 14 DNAshapeR descriptors listed in Table~\ref{tab:dnashape-catalog} with the SantaLucia nearest-neighbor duplex free-energy descriptor defined in Appendix~\ref{app:regional-features}.
Because $\phi(x)_i$ is determined by a $2R+1$ base window and much of that window remains visible when its central nucleotide is masked, the objective can be optimized without integrating evidence beyond a short local context. We therefore compare it with a random 5-mer control, whose target shape and prediction head are matched but whose values carry no biological content. This is the appropriate MFM control because it isolates the biological content of a \emph{local} target while preserving the positional prediction problem. For RPP, random local targets would not test the regional-integration hypothesis; we instead use algorithmic and shuffled-table regional controls that retain regional prediction while removing the biological sequence-to-property mapping.

Table~\ref{tab:mfm-rpp} compares MFM, our local feature-reconstruction objective, with RPP on NTv3-8M.
MFM improves mean AUROC by 1.2 points over the random 5-mer control, but the gain is concentrated in the ClinVar task.
The control is comparable or better on OMIM, gene-expression variants, and CAGE.
Biological RPP reaches 63.8 mean AUROC, exceeding the algorithmic and shuffled-table regional controls by 3.4 and 4.7 points, respectively.
Its largest improvements over both controls occur on ClinVar and TraitGym.
This pattern is consistent with the locality argument of \S\ref{sec:rpp}: local reconstruction can provide auxiliary regularization, 
whereas regional supervision is more effective on tasks that benefit from representations integrating evidence across a broader context.

\begin{table}[h]
\centering
\caption{Local feature reconstruction versus regional property prediction on NTv3-8M.
MFM (bio) uses the composite biological feature provider and MFM (rand) a random 5-mer provider. RPP (alg) predicts generic regional sequence statistics, whereas RPP (shuf) predicts properties recomputed from a shuffled pentamer shape table. Values are AUROC except for CAGE expression, which uses macro Pearson $r$; mean AUROC averages the eight AUROC tasks. Bold marks the best value in each row.}
\label{tab:mfm-rpp}
\scriptsize
\setlength{\tabcolsep}{5pt}
\begin{tabular}{lccccc}
\toprule
\textbf{Task} & \shortstack{\textbf{MFM}\\\textbf{(bio)}} & \shortstack{\textbf{MFM}\\\textbf{(rand)}} & \shortstack{\textbf{RPP}\\\textbf{(alg)}} & \shortstack{\textbf{RPP}\\\textbf{(shuf)}} & \shortstack{\textbf{RPP}\\\textbf{(bio)}} \\
\midrule
Clinical pathogenicity (ClinVar) & \textbf{67.2} & 60.7 & 54.0 & 53.8 & 61.7 \\
Disease variants (OMIM)          & 52.6 & 53.5 & 51.9 & 47.0 & \textbf{59.0} \\
Mendelian variants (TraitGym)    & 49.0 & 48.0 & 48.9 & 47.2 & \textbf{55.8} \\
BRCA1 variant effect             & 51.5 & 49.8 & 50.2 & 48.9 & \textbf{51.6} \\
Transcription-factor binding     & \textbf{81.2} & 81.1 & 78.9 & 78.7 & 80.8 \\
Regulatory variants (eQTL)       & 69.1 & 68.4 & 69.0 & 67.2 & \textbf{70.2} \\
Gene-expression variants         & 62.6 & 63.1 & 63.1 & 63.0 & \textbf{63.3} \\
Chromatin state (histone)        & 67.5 & 66.7 & 67.3 & 67.1 & \textbf{68.1} \\
CAGE expression                 & 33.1 & 33.2 & 32.8 & 29.2 & \textbf{34.9} \\
\midrule
\textbf{Mean AUROC}              & 62.6 & 61.4 & 60.4 & 59.1 & \textbf{63.8} \\
\bottomrule
\end{tabular}
\end{table}

\section{Learned Utilization of the DNAshape Feature Pathway}
\label{app:feature-pathway}

We inspect the trained local-feature-injection pathway of the NTv3-8M VANDAM-MT checkpoint. This analysis uses 256 OpenGenome2 test sequences (2{,}097{,}152 valid A/C/G/T positions).

Let $e_i=\mathbf{e}(x_i)\in\mathbb{R}^D$ be the token embedding and let $f_i=\phi(x)_i$ be the z-scored 14-channel DNAshape vector used in this analysis. The feature projection of \S\ref{sec:lfi} is $h_i=W_f f_i+b_f\in\mathbb{R}^H$. Writing the learned mixing projection as $W_m=[W_m^{\mathrm{seq}}\mid W_m^{\mathrm{feat}}]$, the branch contributions are
$z_i^{\mathrm{seq}}=W_m^{\mathrm{seq}}e_i$ and $z_i^{\mathrm{feat}}=W_m^{\mathrm{feat}}h_i$. To measure position-varying rather than constant contributions, we center each branch over valid positions and report
\begin{equation}
\log_{10}\frac{\left\lVert z_i^{\mathrm{feat}}-\overline{z}^{\mathrm{feat}}\right\rVert_2}
{\left\lVert z_i^{\mathrm{seq}}-\overline{z}^{\mathrm{seq}}\right\rVert_2+\epsilon}.
\end{equation}
Here $\epsilon=10^{-8}$ is a numerical stabilizer. We also compose the feature encoder and feature block of the mixing projection, $W_{\mathrm{eff}}=W_m^{\mathrm{feat}}W_f$. For channel $f$, changing that coordinate by one empirical standard deviation $\sigma_f$ changes the linear mixed embedding by $\sigma_f W_{\mathrm{eff}}[:,f]$ when the remaining coordinates are fixed; we report the magnitude of this mixed-embedding change induced by a one-standard-deviation feature perturbation, $\sigma_f\lVert W_{\mathrm{eff}}[:,f]\rVert_2$.

\begin{figure}[t]
\centering
\includegraphics[width=\textwidth]{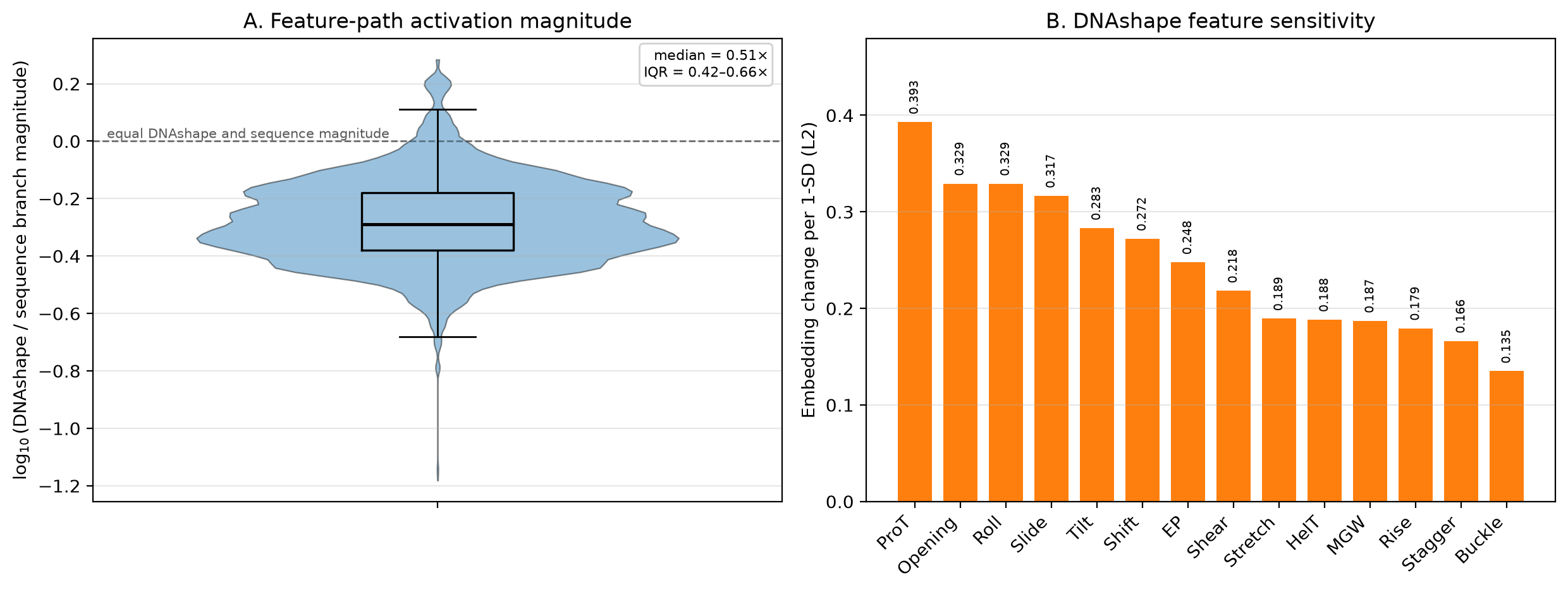}
\caption{Learned utilization of the DNAshape local-feature-injection pathway. \textbf{A:} Distribution across valid positions of the centered DNAshape-branch contribution relative to the centered sequence-embedding contribution after the learned mixing projection. \textbf{B:} Mixed-embedding change induced by a one-standard-deviation perturbation of each feature, computed from the composed feature-encoder and mixer map. These analyses describe learned activation and parameter scales, not causal downstream-task importance.}
\label{fig:feature-pathway}
\end{figure}

Figure~\ref{fig:feature-pathway} shows that the DNAshape branch is not suppressed: its centered contribution has median magnitude $0.51\times$ that of the sequence branch (IQR $0.42$--$0.66\times$). The composed map also reweights channels non-uniformly: ProT induces the largest mixed-embedding change under a one-standard-deviation feature perturbation (0.393), followed by Opening and Roll (0.329 each), whereas Buckle is smallest (0.135). These are descriptive properties of a single trained checkpoint; they show that the injector retains and selectively transforms DNAshape inputs, but do not establish causal importance for an individual feature or downstream task.

\section{RPP Target Ablation}
\label{app:rpp-target-ablation}

We ablate the six RPP targets used by VANDAM-PT on NTv3-8M: intrinsic curvature, AT content, melting propensity, G-quadruplex propensity, cruciform propensity, and Z-DNA propensity. All arms use the same 54{,}000-step continual-pre-training protocol, native MLM objective, 300-bp tiles, scalar-attention pooling, and total RPP coefficient ($\lambda_{\max}=0.5$). Each ablation removes the indicated target or target group while retaining the remaining targets. Table~\ref{tab:rpp-target-ablation} shows that the effects are task-dependent: no target removal lowers every score.

\begin{table}[H]
\centering
\caption{NTv3-8M VANDAM-PT ablation of RPP targets. ``$-$ folding'' removes G-quadruplex, cruciform, and Z-DNA propensities; ``$-$ origin'' removes AT content, melting propensity, and intrinsic curvature; ``$-$ curvature'' removes intrinsic curvature. For supervised tasks, all columns use the same fixed probe seed, so differences isolate the effect of removing each RPP target. Zero-shot scores are seed-independent.}
\label{tab:rpp-target-ablation}
\scriptsize
\setlength{\tabcolsep}{4pt}
\begin{tabular}{lcccc}
\toprule
\textbf{Task} & \textbf{Full RPP} & \textbf{$-$ folding} & \textbf{$-$ origin} & \textbf{$-$ curvature} \\
\midrule
Transcription-factor binding & 80.91 & 80.44 & 80.50 & 80.72 \\
Regulatory variants (eQTL)  & 70.20 & 68.54 & 68.56 & 68.72 \\
Gene-expression variants    & 63.44 & 63.77 & 63.09 & 63.33 \\
BRCA1 variant effect        & 51.55 & 52.49 & 53.02 & 48.87 \\
Disease variants (OMIM)    & 59.00 & 49.72 & 53.28 & 49.51 \\
Mendelian variants (TraitGym) & 55.78 & 56.83 & 56.67 & 52.16 \\
Clinical pathogenicity (ClinVar) & 61.66 & 60.27 & 62.68 & 65.95 \\
Chromatin state (histone)   & 68.04 & 66.94 & 66.91 & 67.10 \\
CAGE expression             & 34.80 & 34.12 & 32.90 & 33.68 \\
\bottomrule
\end{tabular}
\end{table}

The effects are task-dependent: no target removal lowers every score. Nevertheless, removing any target group decreases eQTL, histone, and CAGE performance, while the largest declines occur on OMIM when folding or curvature supervision is removed (59.00 to 49.72 and 49.51, respectively). These results indicate that the regional properties provide complementary signals rather than identifying a single target that is uniformly responsible for the downstream gains.

\section{Intrinsic Token-Objective Evaluation}
\label{app:intrinsic}

We evaluate each checkpoint on the held-out OpenGenome2 test shards using its backbone's native token objective. These metrics complement the downstream evaluations: they ask whether VANDAM changes native-token performance, rather than whether its representations support the held-out functional tasks.

NTv3-8M, Caduceus, and HyenaDNA are evaluated on 3{,}733 deterministic non-overlapping 8{,}192-bp windows; DNABERT-2 is evaluated on 10{,}378 2{,}500-bp windows. For MLM backbones, we use standard random 15\% masking with a fixed seed. 
Loss and accuracy are computed over masked tokens. For causal HyenaDNA, they are computed over all scored next-token positions. 
We report mean cross-entropy loss and top-1 accuracy. Tables~\ref{tab:intrinsic-pt} and~\ref{tab:intrinsic-mt} show that VANDAM's downstream gains do not require uniform improvement in native-token performance.

\begin{table}[H]
\centering
\caption{Intrinsic test metrics after continual pre-training. Each cell gives Baseline-PT $\to$ VANDAM-PT.}
\label{tab:intrinsic-pt}
\scriptsize
\setlength{\tabcolsep}{4pt}
\begin{tabular}{lccc}
\toprule
\textbf{Backbone} & \textbf{Objective} & \textbf{Loss} $\downarrow$ & \textbf{Accuracy} $\uparrow$ \\
\midrule
NTv3-8M   & MLM & $1.098 \to 1.110$ & $50.6\% \to 50.2\%$ \\
Caduceus  & MLM & $1.095 \to 1.109$ & $51.1\% \to 50.2\%$ \\
DNABERT-2 & MLM (BPE) & $5.060 \to 5.058$ & $16.4\% \to 16.4\%$ \\
HyenaDNA  & Causal NTP & $1.265 \to 1.250$ & $40.2\% \to 41.3\%$ \\
\bottomrule
\end{tabular}
\end{table}

\begin{table}[H]
\centering
\caption{Intrinsic test metrics after multi-task training. Each cell gives Baseline-MT $\to$ VANDAM-MT.}
\label{tab:intrinsic-mt}
\scriptsize
\setlength{\tabcolsep}{4pt}
\begin{tabular}{lccc}
\toprule
\textbf{Backbone} & \textbf{Objective} & \textbf{Loss} $\downarrow$ & \textbf{Accuracy} $\uparrow$ \\
\midrule
NTv3-8M   & MLM & $1.136 \to 1.118$ & $48.6\% \to 49.7\%$ \\
Caduceus  & MLM & $1.207 \to 1.195$ & $44.0\% \to 44.8\%$ \\
DNABERT-2 & MLM (BPE) & $5.007 \to 5.673$ & $16.8\% \to 14.3\%$ \\
HyenaDNA  & Causal NTP & $1.267 \to 1.252$ & $40.1\% \to 41.1\%$ \\
\bottomrule
\end{tabular}
\end{table}

\section{Biological-Content Controls}
\label{app:biological-controls}

This section specifies the matched controls used in Figure~\ref{fig:biological-content-controls}. Each control uses the same backbone, data, and training budget as its biological comparison; the intervention is the content of the injected features or RPP targets.

\paragraph{Local-feature-injection control.}
For the LFI (rand) control, the biological local descriptor vector $\phi(x)_i$ is replaced by a random vector of the same dimension at every position. This leaves the injector's parameter count and gradient path unchanged while removing the sequence-derived molecular signal.

\paragraph{RPP target controls.}
The shuffled-table control applies one fixed permutation to the rows of the pentamer shape table and recomputes the same regional properties through the same estimators. It therefore preserves the targets' marginal distributions, cross-feature correlations, and order sensitivity, while destroying the pentamer-to-shape correspondence. The algorithmic control instead predicts three generic regional sequence statistics with no molecular interpretation: regional compression ratio, third-order entropy rate, and minimizer density. Two of these depend on base arrangement rather than composition alone, preserving the regional integration requirement of RPP while removing the biological sequence-to-property mapping.

\paragraph{Multi-task configurations.}
LFI (rand) + RPP retains biological RPP while randomizing the injected descriptors. LFI (bio) + RPP (shuf) and LFI (bio) + RPP (alg) retain biological injected descriptors while replacing only the RPP targets. Thus, each multi-task comparison isolates the content of one biological pathway while holding the remaining architecture and training setup fixed.

\section{qDNA Representation Probing}
\label{app:qdna-probes}

This analysis tests whether the RPP objective changes the linear accessibility of a molecular descriptor family absent from its supervision.
We compare three frozen NTv3-8M checkpoints: the original released checkpoint (Baseline), Baseline-PT, and VANDAM-PT.

\paragraph{Targets and representations.}

qDNA combines a tight-binding description of sequence-dependent electronic couplings with open-quantum-system dynamics to characterize how charge and excitation behave in DNA \citep{herb2025quantumdna}, using electronic parameters from \citet{hawke2010electronic}.
From qDNA simulations, we derive eight molecular descriptors of hole and electron localization and transfer, exciton decay, and electron--hole separation.
They are distinct from the geometry, melting, and non-B-structure properties used to train VANDAM-PT, and no qDNA descriptor enters the RPP loss.
At each nucleotide position, the centered heptamer indexes the corresponding qDNA values.
For descriptor $j$, the qDNA target for a 300-bp tile $t$ is
\begin{equation}
y_{t,j}=\frac{1}{|V_t|}\sum_{i\in V_t}Q_j(x_{i-3},\ldots,x_{i+3}),
\end{equation}
where $V_t$ contains positions with complete canonical centered heptamers.

\paragraph{Residual targets.}
To isolate qDNA variation not linearly predicted by the six regional properties used as RPP targets (Section~\ref{sec:rpp}), we form a six-dimensional nuisance vector $z_t$ from intrinsic curvature, AT content, melting propensity, G-quadruplex propensity, cruciform propensity, and Z-DNA propensity. We standardize these variables using development data only and, separately for each qDNA descriptor, define
\begin{equation}
(\hat{\alpha}_j,\hat{\beta}_j)=\arg\min_{\alpha,\beta}\sum_{t\in\mathcal{D}_{\mathrm{dev}}}\left(y_{t,j}-\alpha-\beta^\top z_t\right)^2,
\qquad
y^{\mathrm{res}}_{t,j}=y_{t,j}-\hat{\alpha}_j-\hat{\beta}_j^\top z_t.
\end{equation}
The fitted standardization and OLS model are applied unchanged to discovery and confirmation data. Thus, residualization removes only the component explained by this linear six-property basis; it does not imply independence from all sequence properties.

The descriptor set comprises central population, maximum transfer, and inverse participation ratio (IPR) separately for holes and electrons, together with exciton lifetime and mean electron--hole separation.
Central population measures the time-averaged population retained at the initial base; maximum transfer is the largest time-averaged transfer away from it; and IPR measures state delocalization.
Exciton lifetime is the time to decay to $1/e$, while electron--hole separation is their time-averaged distance.

We sample 400 OpenGenome2 validation sequences of length 8{,}192 and assign entire source sequences to development (280), discovery (60), or confirmation (60), so tiles from a source cannot cross splits.
Each sequence yields 27 non-overlapping 300-bp tiles, giving 7{,}560 development and 1{,}620 tiles in each held-out phase before any invalid-tile removal.
Every complete sequence is encoded by each frozen backbone, and the final hidden states within a tile are mean-pooled to form its representation.

\paragraph{Probes and inference.}

For every checkpoint--target pair, we fit a ridge-regression probe on development tiles.
The penalty is selected from $\{0.1,1,10,100,1000,10000\}$ using five-fold grouped cross-validation, with folds defined by source sequence and representation standardization fitted within each training fold.
The selected probe is refitted on the full development split and evaluated without refitting on discovery and confirmation.

The primary effect is $\Delta R^2=R^2_{\text{VANDAM-PT}}-R^2_{\text{Baseline-PT}}$.
Confidence intervals use 2{,}000 paired bootstrap replicates over held-out source sequences, retaining all tiles from each sampled source.
One-sided $p$-values use 5{,}000 source-cluster permutations that swap the two checkpoints' fixed predictions within a random subset of sources.
Benjamini--Hochberg correction is applied separately over the eight-descriptor family for the original and residual-target analyses.
A target enters confirmation only when its discovery $q\leq0.05$ and $\Delta R^2\geq0.02$; final selection requires the same two criteria on the independent confirmation split.

\paragraph{Complete results.}

Table~\ref{tab:qdna-probes} gives confirmation performance for all eight reported qDNA descriptors and both target definitions.
The same six descriptors meet both selection stages in the original and residual-target analyses. Their residual-target effects range from $0.051$ to $0.091$ absolute $R^2$ (mean $0.069$), while both maximum-transfer channels remain exceptions.

\begin{table}[H]
\centering
\caption{Frozen linear probing of qDNA descriptors at 300 bp on the independent confirmation split. Checkpoint $R^2$ values use the original targets. Both $\Delta R^2$ columns compare VANDAM-PT with Baseline-PT; intervals are paired source-sequence-bootstrap 95\% confidence intervals.}
\label{tab:qdna-probes}
\scriptsize
\setlength{\tabcolsep}{1.5pt}
\begin{tabular}{lrrrrr}
\toprule
\textbf{Descriptor} & \textbf{Baseline $R^2$} & \textbf{Baseline-PT $R^2$} & \textbf{VANDAM-PT $R^2$} & \textbf{Original $\Delta R^2$ [95\% CI]} & \textbf{Residual $\Delta R^2$ [95\% CI]} \\
\midrule
Hole central population       & 0.717 & 0.709 & 0.769 & $0.060$ [$0.023,0.104$] & $0.054$ [$0.027,0.083$] \\
Hole maximum transfer         & 0.300 & 0.310 & 0.268 & $-0.041$ [$-0.113,0.006$] & $-0.032$ [$-0.078,0.007$] \\
Hole delocalization (IPR)     & 0.766 & 0.742 & 0.812 & $0.071$ [$0.040,0.106$] & $0.091$ [$0.063,0.122$] \\
Electron central population   & 0.671 & 0.663 & 0.746 & $0.083$ [$0.050,0.122$] & $0.057$ [$0.027,0.089$] \\
Electron maximum transfer     & 0.343 & 0.353 & 0.298 & $-0.055$ [$-0.186,0.025$] & $-0.046$ [$-0.194,0.041$] \\
Electron delocalization (IPR) & 0.772 & 0.771 & 0.838 & $0.067$ [$0.047,0.090$] & $0.051$ [$0.021,0.082$] \\
Exciton lifetime              & 0.742 & 0.721 & 0.794 & $0.073$ [$0.053,0.095$] & $0.084$ [$0.053,0.121$] \\
Electron--hole separation     & 0.786 & 0.758 & 0.826 & $0.069$ [$0.049,0.091$] & $0.076$ [$0.052,0.109$] \\
\bottomrule
\end{tabular}
\end{table}
\section{Sensitivity to Masked-Base Handling in LFI}
\label{app:lfi-mask-sensitivity}

The reported LFI policy sets a feature vector to zero whenever its determining pentamer contains an explicit $\langle\mathrm{mask}\rangle$ token, preventing the feature stream from revealing a masked nucleotide. Under the 15\% MLM masking rate and standard 80/10/10 replacement policy, a nucleotide is replaced by $\langle\mathrm{mask}\rangle$ with probability $0.15\times0.8=0.12$. An interior pentamer therefore contains at least one mask with probability $1-(1-0.12)^5\approx47.2\%$, making the injected feature stream sparse during MLM training.

We evaluated two leakage-safe alternatives on NTv3-8M VANDAM-MT. Both construct a separate completed sequence used only to compute DNAshape features; the backbone still receives the original corrupted sequence. \emph{Uniform} replaces each masked base with A, C, G, or T with equal probability. \emph{Conditional} samples it from the OpenGenome2 pentamer distribution conditioned on the visible pentamer context, with pseudocount smoothing. Replacements are resampled during training and applied simultaneously, so adjacent masks introduce no ordering effect. A single replacement can restore features for up to five overlapping pentamers.

\begin{table}[H]
\centering
\caption{Sensitivity of NTv3-8M VANDAM-MT to masked-base handling. Scores are AUROC except CAGE expression, which uses macro Pearson $r$.}
\label{tab:lfi-mask-sensitivity}
\scriptsize
\setlength{\tabcolsep}{2pt}
\begin{tabular}{lccccccccc}
\toprule
\textbf{Handling} & \textbf{ClinVar} & \textbf{OMIM} & \textbf{TraitGym} & \textbf{BRCA1} & \textbf{TF} & \textbf{eQTL} & \shortstack{\textbf{Gene}\\\textbf{expression}} & \textbf{Histone} & \textbf{CAGE} \\
\midrule
Zeroing (VANDAM) & 70.2 & 57.4 & \textbf{56.8} & 54.6 & \textbf{79.8} & 69.3 & 63.1 & 69.3 & \textbf{35.9} \\
Uniform          & 68.6 & \textbf{58.1} & 56.1 & 59.4 & 78.9 & 69.8 & \textbf{70.7} & 71.0 & 33.3 \\
Conditional      & \textbf{70.8} & 55.9 & 54.3 & \textbf{61.2} & 78.6 & \textbf{70.1} & 70.0 & \textbf{71.4} & 33.5 \\
\bottomrule
\end{tabular}
\end{table}

Effects vary by task: imputation improves BRCA1, gene-expression variants, and histone prediction, while both alternatives reduce TF-binding and CAGE performance. Conditional sampling does not consistently outperform uniform sampling. Because this sensitivity analysis is limited to NTv3-8M, we retain zeroing as the reported multi-backbone policy.

\end{document}